\makeatletter
\def\input@path{{packages_eaai/}}
\makeatother

\documentclass[a4paper,fleqn]{cas-sc}

\usepackage{amsmath,amssymb,amsfonts}
\usepackage{algorithmic}
\usepackage{graphicx}
\graphicspath{{final/figs/}{figs/}}
\usepackage{float}
\usepackage[caption=false]{subfig}
\usepackage{textcomp}
\usepackage{multirow}

\usepackage[authoryear]{natbib}
\def\tsc#1{\csdef{#1}{\textsc{\lowercase{#1}}\xspace}}
\tsc{WGM}
\tsc{QE}
\tsc{EP}
\tsc{PMS}
\tsc{BEC}
\tsc{DE}

\usepackage{packages_eaai/irap_acronyms}

\begin{document}
\let\WriteBookmarks\relax
\def\floatpagepagefraction{1}
\def\textpagefraction{.001}

\shorttitle{}    

\shortauthors{}  

\title [mode = title]{Instance-Enriched Semantic Maps for Visual Language Navigation}  

\author[aff1]{Jiho Hong}[orcid=0009-0009-8508-4242]
\fnmark[1]
\author[aff1]{Eunae Kang}[orcid=0009-0009-9019-6396]
\fnmark[1]
\author[aff1,aff2]{Sanghyun Kim}[orcid=0000-0002-0222-8688]
\cormark[1]
\ead{kim87@khu.ac.kr}
\author[aff3]{Young-Sik Shin}[orcid=0000-0002-9653-0633]
\cormark[1]
\ead{yshin86@knu.ac.kr}

\affiliation[aff1]{organization={Department of Mechanical Engineering, Kyung Hee University}, city={Yongin-si}, postcode={17104}, country={Republic of Korea}}
\affiliation[aff2]{organization={Advanced Institutes of Convergence Technology (AICT)}, city={Suwon}, postcode={16229}, country={Republic of Korea}}
\affiliation[aff3]{organization={School of Mechanical Engineering, Kyungpook National University}, city={Daegu}, postcode={41566}, country={Republic of Korea}}

\cortext[1]{Corresponding authors}
\fntext[1]{First authors: These authors contributed equally to this work.}

\nonumnote{This research was supported by the National Research Foundation of Korea (NRF) grant funded by the Korea government (MSIT) (No. RS-2024-00461583, RS-2024-00411007) and by the Korea Basic Science Institute (National Research Facilities and Equipment Center) grant funded by the Ministry of Science and ICT (No. RS-2025-00564593), and National Research Council of Science \& Technology through Project titled “Development of Core Technologies for Robot General Purpose Task Artificial Intelligence Framework” under Grant NK261B.}

\begin{abstract}
Visual Language Navigation (VLN) aims to enable an embodied agent to navigate complex environments by following natural language instructions. Recent approaches build semantic spatial maps and leverage Large Language Models (LLMs) for reasoning and decision making. Despite these advances, existing systems lack instance-level object detail and robustness to diverse user queries, limiting reliable navigation in complex indoor environments. To address these limitations, we propose Instance-Enriched Semantic Maps, a unified framework with three key contributions: (1) Instance-level two-and-a-half-dimensional (2.5D) rich information mapping that constructs maps from color and depth observations via open-vocabulary panoptic segmentation, preserving vertical distinctions and capturing small objects, while storing diverse semantic attributes and natural language captions enriched with room-level context. (2) Robust query processing via LLM-based target selection, which dynamically routes queries across type-specialized experts and integrates their outputs through score-level fusion, enabling consistent goal selection across diverse query formulations. (3) Storage-efficient semantic representation that achieves approximately 96\% reduction compared to three-dimensional (3D) scene-graph approaches while preserving sufficient spatial information for navigation. The proposed 2.5D representation outperforms the 3D baseline by over 27\% in prediction-normalized Area Under the Curve (AUC). In navigation experiments, our method achieves over 17\% improvement in object retrieval and over 23\% in navigation success compared to the baseline across diverse query types. The project page is available at \url{https://rcilab.github.io/iesm_vln}.
\end{abstract}



\maketitle

\section{Introduction}
\label{sec:introduction}

\begin{figure}[t]
    \centering
    \includegraphics[width=\textwidth]{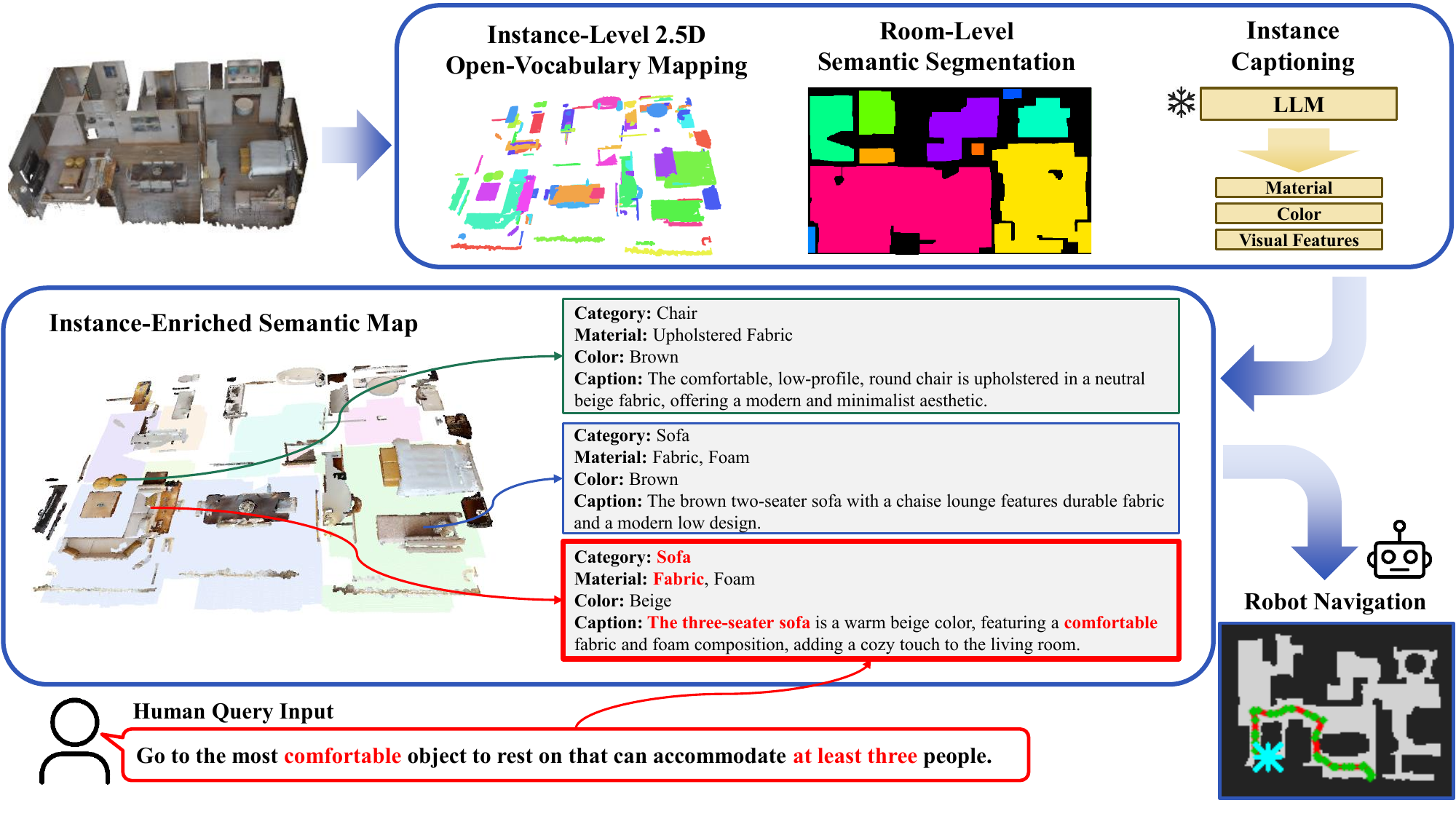}
    \caption{Instance-Enriched Semantic Map is an open-vocabulary 2.5D map representation built upon Instance-Level 2.5D Open-Vocabulary Mapping, Room-Level Semantic Segmentation, and Instance Captioning. During navigation, the Multi-Type Expert Fusion Retrieval module interprets diverse natural language queries by routing inference across type-specialized experts and fusing their outputs, enabling robust zero-shot goal instance selection from the enriched map.}
    \label{fig1}
\end{figure}

Achieving robust operation in complex and dynamic environments remains a fundamental challenge for autonomous robots \citep{2024robotapp, 2024realhl, 2025astar, 2022eaaisurvey}. A key step toward overcoming this challenge lies in enabling robots to understand and act upon human language, which allows them to interpret high-level intent and adapt their behavior accordingly \citep{2017interpret, 2023vlmaps, 2024eaaipasts}. 
In this context, \ac{VLN} has emerged as a promising approach that enables embodied agents to follow natural language instructions in photo-realistic environments \citep{anderson2018vln, 2019habitat, 2017ai2thor}. 

A core design challenge in \ac{VLN} lies in how to represent the environment so that agents can jointly perform spatial reasoning and language grounding. Recent approaches leverage pre-trained \ac{VLMs} to construct semantic maps and achieve zero-shot navigation without task-specific fine-tuning, demonstrating the growing potential of open vocabulary understanding in embodied agents \citep{2023vlmaps, 2023cf, 2024hovsg}. However, effectively integrating open-vocabulary reasoning with spatial representations that balance semantic expressiveness and storage efficiency remains challenging.

Early work typically relied on graph-structured panoramas \citep{anderson2018vln, fried2018speaker} or \ac{2D} feature maps. While these representations offer foundational semantic cues, they inherently lack instance-level granularity and vertical structural awareness. Consequently, such systems often struggle to disambiguate multiple instances within the same semantic category or interpret nuanced, attribute-centric instructions. To address these geometric constraints, recent research has pivoted toward high-fidelity \ac{3D} representations, including hierarchical scene graphs \citep{2023cf, yin2024sg}, volumetric reconstructions \citep{2024ver, 2025g3d}, and Gaussian-based 3D splatting \citep{2024unitedvln, 2025gaussnav}. Although these paradigms facilitate superior spatial-language reasoning, they introduce substantial storage requirements, which limits their scalability in large-scale or long-term robotic deployments.

In this work, we propose Instance-Enriched Semantic Maps, a unified framework for zero-shot open-vocabulary \ac{VLN} that integrates instance-level semantic detail with a compact \ac{2.5D} spatial representation. This framework bridges the gap between the expressiveness of 3D methods and the efficiency of \ac{2D} approaches by preserving vertical distinctions through multi-occupancy grid cells while maintaining storage costs well below those of full 3D representations. The framework enriches each instance with room-level context and descriptive captions covering visual properties such as color and material, enabling fine-grained semantic understanding beyond categorical labels. On top of this enriched representation, the proposed \ac{MTEFR} module enables robust goal instance selection across structurally diverse natural language queries through query-adaptive expert routing and score-level fusion. As illustrated in Fig. \ref{fig1}, the framework comprises three mapping components and one \ac{MTEFR}-based navigation component, detailed in Sec. \ref{sec:method}.

The main contributions of this paper are as follows:
\begin{itemize}
    \item \textbf{Instance-level 2.5D rich information mapping}: A comprehensive mapping framework is introduced that stores diverse semantic attributes at the instance level, such as spatial context, visual properties, descriptive captions, and room-level context, thereby significantly improving navigation accuracy through enhanced object understanding and disambiguation. The 2.5D representation further ensures that even small objects are reliably preserved without being lost during the mapping process.
    \item \textbf{Robust query processing via \ac{LLM}-based target selection}: The proposed \ac{MTEFR} module actively leverages the diverse semantic attributes stored at the instance level through query-adaptive expert routing and score-level fusion, achieving consistent and robust goal instance selection across structurally diverse query formulations, moving beyond conventional keyword extraction to fusion-based semantic reasoning.
    \item \textbf{Storage-efficient semantic representation}: A 2.5D representation that combines instance-centric mapping with \ac{LLM}-based captioning significantly reduces storage requirements compared to 3D scene-graph approaches while preserving sufficient spatial information for precise navigation.
\end{itemize}

Extensive experiments on the \ac{MP3D}, Replica, and \ac{HM3DSEM} datasets validate each contribution. In Sec. \ref{subsec:exp1}, the proposed 2.5D mapping achieves accurate instance-level segmentation, with the prediction-normalized \ac{AUC} achieving an improvement of over 27\% over the 3D baseline on average across datasets, including reliable preservation of small objects. For room-level segmentation in Sec. \ref{subsec:exp2}, our method effectively suppresses pixel-level noise by leveraging structural priors, achieving an improvement of 14\% in \ac{mAcc} and 22\% in \ac{mIoU} over the \ac{CLIP}-based baseline. In Sec. \ref{subsec:exp3}, our method achieves an average improvement of over 17\% in object retrieval and 23\% in navigation success given correct object retrieval across various query types compared to \ac{HOV-SG} \citep{2024hovsg}, with the largest gains observed on description-based and abstract queries, and ablation studies confirming the contribution of each module including \ac{MTEFR}. Sec. \ref{subsec:exp4} demonstrates approximately 96\% storage reduction compared to 3D scene-graph approaches.

The remainder of this paper is organized as follows. Sec.\ref{sec:related_work} reviews related work. Sec. \ref{sec:method} details the instance-enriched mapping framework, including 2.5D map construction, room-level segmentation, descriptive captioning, and \ac{MTEFR}-based goal-directed navigation. Sec. \ref{sec:exp} presents experimental results in simulated environments. Sec. \ref{sec:conclusion} concludes with a summary and future directions.
\section{Related Work}
\label{sec:related_work}
\subsection{Spatial Representations for VLN}
The spatial representation adopted by a \ac{VLN} system determines what geometric and semantic structure is available for grounding language instructions. Existing representations range from flat 2D projections to full 3D reconstructions, each reflecting a different trade-off between storage efficiency and spatial expressiveness.

2D representations \citep{2023bird, 2023bevbert, 2023gridmm} project high-dimensional visual features onto a flat plane and store them within discrete grid cells or structured graph nodes. While this representation is computationally compact and facilitates efficient spatial queries, it inherently collapses the vertical dimension of the 3D environment. Consequently, when multiple distinct objects occupy the same horizontal footprint at varying heights, such as items placed on vertically stacked shelves, their semantic features are irreversibly blended into a single cell, leading to severe semantic ambiguity during goal retrieval. Recent instance-aware 2D extensions \citep{2023instance, 2024ivlmap} have attempted to mitigate this by maintaining unique object identities. However, they remain fundamentally limited in their inability to resolve vertical overlap, as the underlying 2D projection cannot differentiate between instances stacked along the Z-axis.

To preserve vertical structure, recent methods extend spatial representations to higher dimensions. Volumetric approaches such as \ac{VER} \citep{2024ver} voxelize the environment into structured 3D cells and jointly predict occupancy, room layout, and bounding boxes. VER builds volumetric representations online during navigation for volume state estimation and episodic memory, rather than using only transient per-step local voxels. Implicit feature-field methods such as g3D-LF \citep{2025g3d} pre-train generalizable 3D-language representations via volume rendering and contrastive learning. While generalizable across unseen scenes, these representations lack an explicit instance dictionary. There is no mechanism to associate discrete object identities, room labels, or descriptive captions with individual entities, limiting the ability to handle fine-grained attribute-based queries. Neural rendering approaches \citep{2024unitedvln, 20253dgaussian} adopt 3D Gaussian Splatting \citep{2023gs} to render both appearance and semantic features, but require per-scene optimization or rendering-specific pre-training, tightly coupling the representation to a particular pipeline. At the zero-shot 3D level, scene-graph methods such as \ac{HOV-SG} \citep{2024hovsg} organize objects into structured hierarchies with rich spatial relationships \citep{2023cf,2024cg}. Similarly, OpenMap \citep{2025openmap} builds zero-shot open-vocabulary 3D instance maps via structural-semantic consensus and \ac{LLM} instruction grounding. Like HOV-SG, OpenMap also incurs substantial storage costs as its 3D instance representations scale with scene complexity. These methods provide the richest spatial fidelity, but their storage requirements grow with scene complexity, which limits scalability in large-scale or long-term deployments.

2D representations are efficient but lose vertical information, while 3D methods are expressive but storage-heavy. These complementary limitations motivate the proposed 2.5D instance-level representation, which allows multiple instance indices per grid cell to preserve vertical distinctions in a persistent global map without incurring the storage overhead of full 3D reconstruction.

\subsection{Open-Vocabulary Semantic Mapping}
The core challenge in semantic mapping for \ac{VLN} lies in how environmental observations are grounded into linguistic concepts and organized within a spatial structure. A primary distinction in this domain is whether the mapping system relies on task-specific training or leverages the zero-shot capabilities of pre-trained \ac{VLMs}.

\ac{VLMs} pre-trained on large-scale data enable robots to ground natural language instructions in visual observations without fine-tuning \citep{2023distill}, supporting open-vocabulary object recognition and integration into exploration-based navigation frameworks \citep{2023cows,2022lmnav}. Recent works propose zero-shot techniques for semantic spatial maps by integrating visual landmarks with language \citep{2023vlmaps,2023nlmap}. Learning-based \ac{VLN} methods typically optimize navigation policies and map representations end-to-end on task-specific datasets \citep{2023bevbert, 2023gridmm}. These approaches use structured maps as intermediate features for action prediction, demonstrating effectiveness on established benchmarks. However, these methods are inherently bounded by their training distributions and often struggle with generalization to novel environments, exhibiting limited robustness to open-ended, out-of-distribution language queries.

In contrast, zero-shot approaches construct semantic maps entirely by utilizing pre-trained \ac{VLMs} without environment-specific fine-tuning \citep{2023vlmaps, 2024vlngame}. In these frameworks, the choice of \ac{VLM} directly dictates the semantic granularity of the stored information. These approaches typically construct semantic maps from \ac{RGB-D} observations and camera poses acquired during an exploration phase, and subsequently leverage these maps for language-conditioned navigation, a setting our framework likewise adopts. \ac{VLMaps} \citep{2023vlmaps} employs \ac{LSeg} \citep{2022lseg} for dense per-pixel feature projection. While efficient, this pixel-level storage lacks object-level awareness, often leading to the semantic "blending" of distinct objects that occupy the same spatial grid. To achieve better object-level consistency, many recent works have adopted the \ac{SAM}+\ac{CLIP} \citep{2023sam, 2021CLIP} pipeline. However, this combination frequently suffers from over-segmentation and background noise, as \ac{SAM} may generate semantically fragmented regions while \ac{CLIP} extracts features from unmasked, noisy global images. HOV-SG inherits these fragmentation issues due to its reliance on the \ac{SAM}-based pipeline.

Building on these observations, our method adopts a zero-shot instance-level mapping strategy. By utilizing \ac{SEEM} \citep{2023seem}, we produce consistent instance masks that alleviate fragmentation, allowing the system to maintain stable object-level identity across multiple frames. Despite this improvement in segmentation, most existing zero-shot maps still represent instances primarily through fixed-dimensional embeddings, which are often insufficient for resolving attribute-based or abstract queries. To overcome this limitation, we introduce an instance captioning component that augments each instance entry with natural language descriptions to enable richer per-instance semantic content beyond category labels.

\subsection{LLM-based Query Reasoning for Navigation}
\ac{LLMs} have been increasingly adopted to decompose high-level instructions into actionable subgoals for embodied navigation \citep{2022saycan, 2023palme, 2025opennav}. However, these methods primarily focus on instruction decomposition and action planning, lacking instance-level semantic grounding necessary for distinguishing between multiple objects of the same category or handling queries that reference specific visual properties.

Within the zero-shot \ac{VLN} paradigm, \ac{LLMs} serve as the reasoning bridge between user queries and the semantic map. Existing approaches typically support only category-level or attribute-level matching. When the map stores only category labels, the \ac{LLM} cannot distinguish between multiple instances of the same class or interpret queries that involve specific visual properties. Keyword-based query processing, as used in HOV-SG, extracts object and room keywords from the query while discarding descriptive content entirely, so that queries with and without descriptions yield identical retrieval results. This design fundamentally cannot handle abstract references that omit explicit category names, such as "find the most comfortable object to rest on."

These limitations suggest that enriching the per-instance information available to the \ac{LLM} is critical for handling diverse query forms. Motivated by this insight, the proposed framework augments each instance entry with natural language captions that capture detailed visual properties such as color, material, texture, and functional characteristics alongside category labels and room context. Furthermore, rather than passing all instance information uniformly to a single \ac{LLM}, the proposed \ac{MTEFR} module dynamically routes each query across type-specialized experts and integrates their outputs via score-level fusion, enabling fusion-based reasoning that adapts to the informational structure of diverse query formulations.
The proposed framework is detailed in Sec. \ref{sec:method}.

\begin{figure}[!t]
    \centering
    \includegraphics[width=\textwidth]{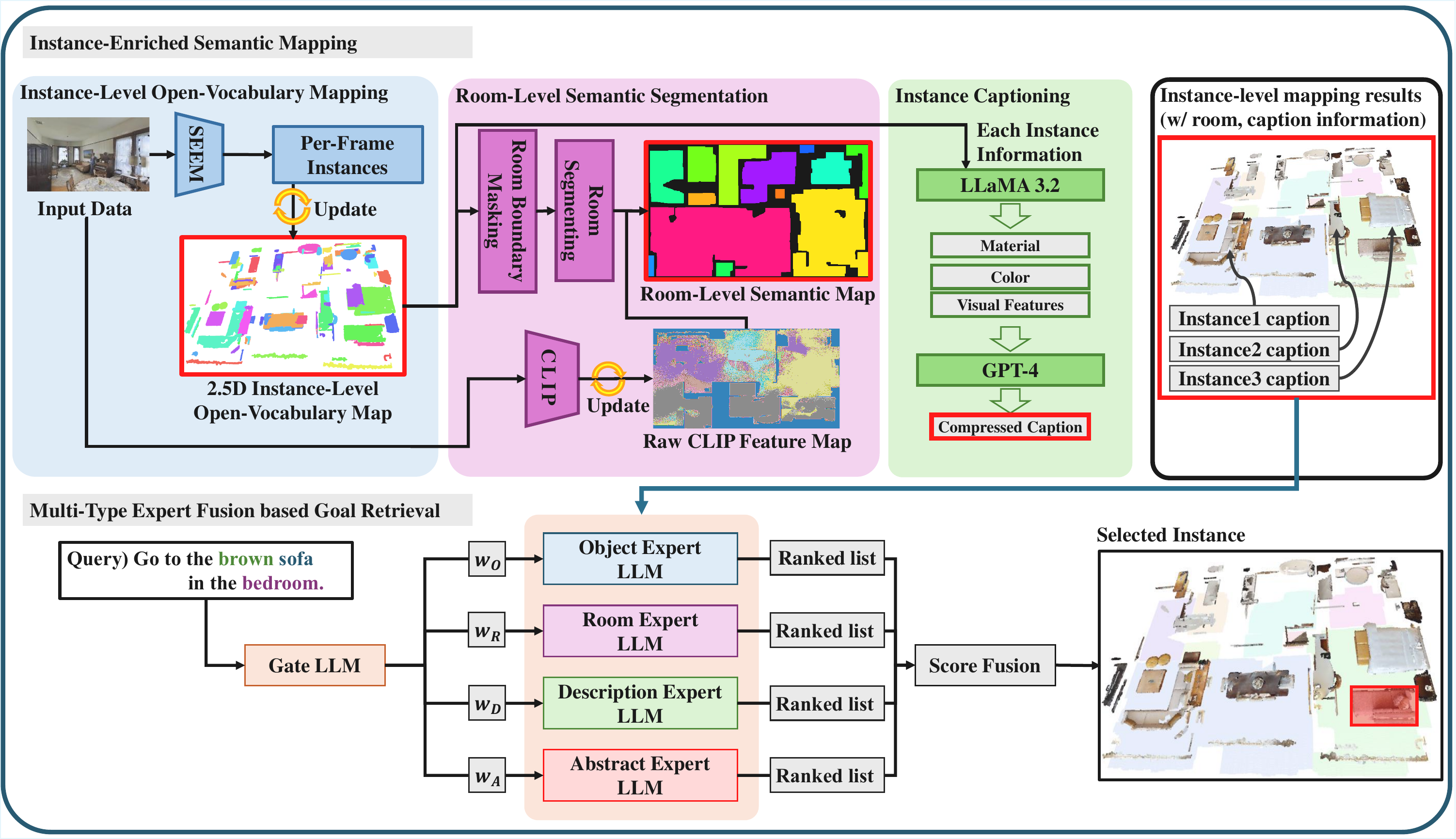}
    \caption{System overview. The proposed framework consists of four main components. The upper section covers Instance-Enriched Semantic Mapping, comprising three stages. First, Instance-Level 2.5D Open-Vocabulary Mapping (left, Sec. \ref{subsec:methodA}) extracts per-frame instance masks and embeddings using \ac{SEEM} and incrementally fuses them into a global 2.5D semantic map in which each grid cell stores indices of multiple vertically stacked instances. Second, Room-Level Semantic Segmentation (center, Sec. \ref{subsec:methodB}) combines structural room boundary masks derived from instance segmentation with \ac{CLIP}-based vision-language similarity to produce a region-consistent semantic room map. Third, Instance Captioning (right, Sec. \ref{subsec:methodC}) processes each instance through LLaMA 3.2 Vision to extract visual attributes including material, color, and appearance, followed by GPT-4 aggregation into a compressed caption, yielding instance-level mapping results enriched with room and caption information. The lower section presents Multi-Type Expert Fusion based Goal Retrieval (Sec. \ref{subsec:methodD}), where a gate LLM assigns query-adaptive weights ($\omega_O,\omega_R,\omega_D,\omega_A$) over four type-specialized expert LLMs, namely the Object, Room, Description, and Abstract experts, each producing an independent ranked list of candidate instances, which are then integrated via score-level fusion to identify the final goal instance.}
    \label{fig2}
\end{figure}

\section{Methodology}
\label{sec:method}
A multi-stage pipeline generates a semantic, instance-centric map from \ac{RGB-D} inputs to support language-based navigation. First, instance segmentation masks are derived from \ac{RGB} inputs using a VLM and merged across frames to create a temporally consistent instance-level 2.5D map. Second, architectural structures are identified from the merged instances to generate a room separation mask, which is combined with an obstacle map to assign room labels via vision-language embeddings, yielding a semantic room map. Third, each object instance is described with captions generated from visual features aggregated across multiple frames, enriching the map with open-vocabulary semantic attributes. Finally, during navigation, the constructed instance-enriched map is leveraged by the proposed \ac{MTEFR} module, which dynamically interprets the user query to route inference across specialized experts and fuses their outputs for robust goal instance selection across diverse query formulations. An overview of the full pipeline is shown in Fig. \ref{fig2}.

\subsection{Instance-Level 2.5D Open-Vocabulary Mapping}
\label{subsec:methodA}
Instance-level open-vocabulary information is mapped into a global 2.5D representation through a five-step pipeline. Candidate masks and embeddings are extracted from \ac{RGB-D} frames, associated with prior instances by geometric and semantic similarity, and fused into consistent entries. Our instance association and fusion process assumes accurate and globally consistent camera poses, obtained either from the simulator or from a reliable \ac{SLAM} estimation. The final output consists of (i) a 2.5D map $\mathcal{M}$ and (ii) an instance dictionary $\mathcal{I}$, detailed in the following steps.

\subsubsection{Instance-Centric Representation}
From a sequence of \ac{RGB-D} observations, the panoptic segmentation module of SEEM extracts candidate instance masks together with their open-vocabulary feature embeddings at each time step. Each candidate instance is reconstructed in 3D from depth data, transformed into the global coordinate frame using the camera pose, and projected top-down onto a grid map.

At time step $t$, the candidate instances are organized into the following dictionary:
\begin{align}
\mathcal{I}^{\text{cand}}_t
  = \left\{\, a \mapsto \mathbf{I}^{\text{cand}}_{t,a} \,\right\}_{a=1,\dots,A_t},
\end{align}
where each entry is defined as
\begin{align}
\mathbf{I}^{\text{cand}}_{t,a}
  = \bigl(\mathbf{m}^{\text{cand}}_{t,a},\, \mathbf{f}^{\text{cand}}_{t,a}\bigr).
\end{align}
\noindent
Here, $\mathbf{m}^{\text{cand}}_{t,a}\in\{0,1\}^{H\times W}$ denotes the top-down binary mask of candidate $a$, and $\mathbf{f}^{\text{cand}}_{t,a}\in\mathbb{R}^{D_S}$ represents its feature embedding.
$D_S$ denotes the dimensionality of SEEM feature embeddings, $A_t$ denotes the number of candidate instances obtained from the $t$-th image frame, and $H, W$ denote the height and width of the top-down grid, respectively.

\subsubsection{Instance Association}
Newly detected candidate instances are compared against the $N_{t-1}$ instances accumulated up to time step $t-1$. The information obtained previously is organized into the following dictionary:
\begin{align}
\mathcal{I}^{\text{exist}}_{t-1}= \left\{\, n \mapsto \mathbf{I}^{\text{exist}}_{t-1,n}\right\}_{n=1,\dots,N_{t-1}} \quad.
\end{align}
Unlike candidate entries, existing instances accumulate information across multiple frames, and therefore each entry additionally maintains attributes that are progressively updated, including the frame index list, mean height, and observation count. Each entry is defined as
\begin{align}
\mathbf{I}^{\text{exist}}_{t-1,n}= \big(\mathbf{m}^{\text{exist}}_{t-1,n},\, \mathbf{f}^{\text{exist}}_{t-1,n},\, \mathbf{g}^{\text{exist}}_{t-1,n},\, \bar{h}^{\text{exist}}_{t-1,n},w^{\text{exist}}_{t-1,n}\big). 
\end{align}
\noindent
Here, $\mathbf{m}^{\text{exist}}_{t-1,n}\in\{0,1\}^{H\times W}$ denotes the 2D top-down mask of instance $n$, and $\mathbf{f}^{\text{exist}}_{t-1,n}\in\mathbb{R}^{D_S}$ is its aggregated feature embedding. $\mathbf{g}^{\text{exist}}_{t-1,n}=[g^{\text{exist}}_{t-1,n,1},\dots,g^{\text{exist}}_{t-1,n,V}]$ is the frame-index list of instance $n$, where $V$ is the total number of frames in which the instance appears, and $g^{\text{exist}}_{t-1,n,v}$ denotes a specific frame index. $\bar{h}^{\text{exist}}_{t-1,n}$ represents the mean height of instance $n$, computed over its occupied grid cells, and  $w^{\text{exist}}_{t-1,n}\in\mathbb{N}$ denotes the weight corresponding to the number of frames in which the instance has been observed up to time step $t-1$.

Matching a newly observed candidate instance $\mathbf{I}^{\text{cand}}_{t,a}$ to an existing instance $\mathbf{I}^{\text{exist}}_{t-1,n}$ relies on geometric and semantic similarities. The geometric similarity $\phi_{\text{geo}}$ is defined as
\begin{equation}
    \phi_{\text{geo}}\bigl(\mathbf{m}_{\alpha},\mathbf{m}_{\beta}\bigr) 
    = \max\!\Bigl(
       \operatorname{o}(\mathbf{m}_{\alpha},\mathbf{m}_{\beta}),
       \operatorname{o}(\mathbf{m}_{\beta},\mathbf{m}_{\alpha})
     \Bigr),
    \label{eq:phi_geo}
\end{equation}
where $\operatorname{o}(\mathbf{m}_{\alpha},\mathbf{m}_{\beta})$ is the overlap ratio, which is the intersection area normalized by the area of $\mathbf{m}_{\alpha}$. Compared with \ac{IoU}, max-overlap mitigates the effects of single-view truncation and top-down compression, allowing matches to be correctly identified even when the overlap is below a predefined threshold. The semantic similarity $\phi_{\text{sem}}$ is given by
\begin{equation}
\phi_{\text{sem}}\bigl(\mathbf{f}_{\alpha},\mathbf{f}_{\beta}\bigr)
  =\mathbf{f}_{\alpha}\,\mathbf{f}_{\beta}^{\top},
\end{equation}
which equals the cosine similarity. The overall similarity is defined as a convex combination:
\begin{equation}
\phi=\omega_{\text{geo}}\phi_{\text{geo}}+
     \omega_{\text{sem}}\phi_{\text{sem}},\qquad
\omega_{\text{geo}}+\omega_{\text{sem}}=1.
\end{equation}

For each candidate entry $\mathbf{I}^{\text{cand}}_{t,a}$, the matched instance index $i$ is selected from the $N_{t-1}$ accumulated entries in $\mathcal{I}^{\text{exist}}_{t-1}$ that maximizes $\phi$. The selected instance satisfies both thresholds $\phi_{\text{geo}}>\delta_{\text{geo}}$ and
$\phi_{\text{sem}}>\delta_{\text{sem}}$:
\begin{equation}
i=\operatorname*{argmax}_{n=1,\dots,N_{t-1}}\phi\bigl(\mathbf{I}^{\text{cand}}_{t,a},\mathbf{I}^{\text{exist}}_{t-1,n}\bigr).
\end{equation}

\subsubsection{Instance Fusion}
Let $\Omega=\{1,\dots,H\} \times \{1,\dots,W\}$ denote the discrete grid domain of the top-down map. If a candidate instance $\mathbf{I}^{\text{cand}}_{t,a}$ does not match any existing entry, it is added as a new detection to the dictionary $\mathcal{I}^{\text{exist}}_{t}$. Otherwise, matched entries are fused as follows. The mask is updated by applying a logical OR between the candidate mask and the existing mask:
\begin{align}
\mathbf{m}^{\text{exist}}_{t,i}(x,y)=\mathbf{m}^{\text{cand}}_{t,a}(x,y) \lor \mathbf{m}^{\text{exist}}_{t-1,i}(x,y), \;\;(x,y)\in\Omega.
\end{align}
\noindent
The feature embedding is updated via a weighted average of
the previous embedding and the candidate embedding, where the previous weight is used as the factor. Once a match is confirmed,
the weight is
incremented by one:
\begin{equation}
\mathbf{f}^{\text{exist}}_{t,i}
= \frac{w^{\text{exist}}_{t-1,i}\,\mathbf{f}^{\text{exist}}_{t-1,i}
       + \mathbf{f}^{\text{cand}}_{t,a}}
       {w^{\text{exist}}_{t-1,i}+1}, \quad
w^{\text{exist}}_{t,i}=w^{\text{exist}}_{t-1,i}+1.
\end{equation}
\noindent
The frame-index list is extended by appending the current frame index, and its items are sorted by occupied pixel area. The mean height is updated using the same weighted averaging scheme as the feature embedding.

After all frames are processed, the dictionary $\mathcal{I}^{\text{exist}}_{T}$ is obtained, where $T$ denotes the total number of time steps. A second association--fusion pass, using the same procedure as above but applied within $\mathcal{I}^{\text{exist}}_{T}$, merges long-range duplicates arising from occlusion, pose drift, or view changes and enforces global instance consistency. This yields the definitive dictionary $\mathcal{I}^{\text{final}}=\{\, n \mapsto \mathbf{I}^{\text{final}}_{n} \,\}_{n=1,\dots,N}$, where $N$ denotes the number of final entries.

\subsubsection{Instance-Level 2.5D Map Creation}
Per-instance masks are aggregated to construct the global 2.5D map:
\begin{equation}
\mathcal{M}=\{\,\mathcal{M}_{x,y}\mid (x,y)\in\Omega\,\}, \quad
\mathcal{M}_{x,y} = [n_1,\dots,n_{K_{x,y}}],
\end{equation}
\noindent
where each cell $\mathcal{M}_{x,y}$ stores the indices of all instances occupying that cell, and $K_{x,y}$ denotes the number of occupying instances at $\mathcal{M}_{x,y}$. 

This multi-occupancy per cell motivates the 2.5D designation, 
as multiple instances can be stacked within a single grid cell. After the map is constructed, the per-instance mask $\mathbf{m}^{\text{final}}_n$ contained in each $\mathbf{I}^{\text{final}}_n$ is discarded to avoid redundancy.

\subsubsection{Open-Vocabulary Instance Categorization}
Finally, instance feature embeddings are categorized using an open‑vocabulary approach. Given a language‑category list
$\mathcal{L}^{\text{icat}}=[\mathbf{l}^{\text{icat}}_{1},\mathbf{l}^{\text{icat}}_{2},\dots,\mathbf{l}^{\text{icat}}_{J}]$, where $\mathbf{l}^{\text{icat}}_{j}$ denotes the $j$-th category text and $J$ is the number of categories, textual embeddings
$[\mathbf{e}^{\text{icat}}_{1},\mathbf{e}^{\text{icat}}_{2}, \dots,\mathbf{e}^{\text{icat}}_{J}],\;\mathbf{e}^{\text{icat}}_{j}\!\in\!\mathbb{R}^{D_S}$ are generated with a pre‑trained SEEM text encoder.
These embeddings constitute the matrix $\mathbf{E}^{\text{icat}}\!\in\!\mathbb{R}^{J\times D_S}$, whose rows correspond to category embeddings. In parallel, embeddings $[\mathbf{f}^{\text{final}}_{1},\dots,\mathbf{f}^{\text{final}}_{N}]$  from $\mathcal{I}^{\text{final}}$ form the matrix $\mathbf{F}^{\text{final}}\!\in\!\mathbb{R}^{N\times D_S}$, 
where each row represents the embedding of an instance.

An instance‑to‑category similarity matrix is then computed as $\mathbf{Z}=\mathbf{F}^{\text{final}}(\mathbf{E}^{\text{icat}})^{\top}$, where $\mathbf{Z}\!\in\!\mathbb{R}^{N\times J}$ stores similarity scores for
all pairs. The category index for instance $n$ is obtained by
\begin{equation}
c_{n}=\operatorname*{argmax}_{j=1,\dots, J}\mathbf{Z}_{nj},
\end{equation}
where $c_{n}$ indicates the position of the selected category in
$\mathcal{L}^{\text{icat}}$.
Using this index, the corresponding category text is retrieved and stored as the label $\mathbf{l}^{\text{final}}_n$ within the definitive entry $\mathbf{I}^{\text{final}}_n$.

Through this pipeline, the instance-level 2.5D open-vocabulary mapping is completed, yielding two final outputs: the global 2.5D map $\mathcal{M}=\{\mathcal{M}_{x,y}\;|\;(x,y)\in\Omega\}$ and the definitive instance dictionary $\mathcal{I}^{\text{final}}=\{n\mapsto \mathbf{I}^{\text{final}}_n\}_{n=1,\dots,N}$, with each entry defined as $\mathbf{I}^{\text{final}}_n=(\mathbf{f}^{\text{final}}_n,\mathbf{g}^{\text{final}}_n,\bar{h}^{\text{final}}_n,w^{\text{final}}_n, \mathbf{l}^{\text{final}}_n)$.

\subsection{Room-Level Semantic Segmentation}
\label{subsec:methodB}
This section presents a two-step approach to generate room segmentation maps. First, individual room instances are separated using structural object masks. Then, a semantic mask is generated by computing the similarity between \ac{CLIP} visual and linguistic features. This mask is combined with the structural map for region-wise aggregation.

\subsubsection{Instance-Level Room Separation}
The instance segmentation masks produced by SEEM serve as the basis for extracting structural elements that typically delineate room boundaries such as walls, doors, and large fixed furniture. To ensure that segmentation occurs only within navigable areas, an obstacle map is incorporated to exclude non-traversable regions. This obstacle map is a binary representation of the environment that marks non-traversable regions, based on depth sensing and occupancy estimation, and its construction is explicitly described in Sec. \ref{subsec:methodD}. The selected masks are merged into a room-boundary mask $\mathbf{m}^\text{room}$, a binary representation in which boundary pixels are set to 1 and all other pixels are set to 0. This binary format enhances structural contrast and serves as an input for subsequent line detection.

The \ac{LSD} \citep{2012lsd} is applied to extract wall-like linear structures. For each pixel $(x,y)$ of the boundary mask $\mathbf{m}^\text{room}$, the horizontal and vertical gradients are computed as
\begin{align}
    z_x(x,y) &= \frac{\mathbf{m}^\text{room}(x+1, y)+\mathbf{m}^\text{room}(x+1, y+1) - \mathbf{m}^\text{room}(x, y)-\mathbf{m}^\text{room}(x, y+1)}{2}, \\
    z_y(x,y) &= \frac{\mathbf{m}^\text{room}(x, y+1)+\mathbf{m}^\text{room}(x+1, y+1) - \mathbf{m}^\text{room}(x, y)- \mathbf{m}^\text{room}(x+1, y)}{2}.
\end{align}

The level-line angle $\theta_{xy}$, perpendicular to the gradient orientation, is then calculated as
\begin{equation}
\theta_{xy}=\arctan\left(\frac{z_x}{-z_y}\right).
\end{equation}
Then, pixels whose level-line angles differ by less than a threshold $\tau$ from a region angle $\theta_\text{region}$, which is initialized to the level-line angle of the seed pixel and progressively updated as pixels are added to the region, are grouped into connected regions, referred to as \ac{LSRs}. A pixel $(x,y)$ belongs to the $\mathcal{S}^\text{cand}=\{{\mathbf{s}^\text{cand}_c}\}_{c=1\ldots C}$, where $C$ denotes the number of \ac{LSRs} if $\left|\theta_{xy}-\theta_\text{region}\right|<\tau$.
Each LSR $\mathbf{s}^\text{cand}_c$ is approximated by a minimum bounding rectangle $\mathbf{r}_{c}$. For each candidate rectangle, the total number of pixels $q$ and the number of aligned points $o$ are counted.
To reject false detections, the \ac{NFA} metric is computed as
\begin{equation}
\text{NFA}(\mathbf{r}_c) = (HW)^{5/2}\cdot\eta\cdot \sum_{j = o}^{q} \binom{q}{j} \, p^{\,j} \, (1-p)^{\,q-j},
\end{equation}
where $p=\tau/\pi$ denotes the probability that a pixel is aligned with the rectangle orientation under the a contrario noise model, in which level-line angles are independently and uniformly distributed over $[0,2\pi]$. The term $(HW)^{5/2}$ accounts for the total number of possible oriented rectangles in the $H\times W$ grid, and $\eta$ denotes the number of distinct precision values tested. Only line segments satisfying $\text{NFA}\leq 1$ are retained.

Detected line segments shorter than a length threshold are discarded, and adjacent collinear segments are merged to improve continuity. The accepted line segments are rasterized into a boundary mask $\mathbf{m}^\text{room}$, which is refined via morphological dilation and closing operations to fill gaps and connect fragmented segments. Finally, connected component analysis is performed on the refined mask to extract individual room instances. Each component is assigned a unique instance ID, resulting in a structural room instance map $\mathcal{S}^\text{final}=\{{\mathbf{s}^\text{final}_b}\}_{b=1\ldots B}$, where $B$ denotes the number of room masks. While this map delineates individual rooms, it lacks semantic information about room types, thereby requiring a second stage for semantic labeling.

\subsubsection{Semantic Labeling}

Dense image-text embedding maps are extracted using the \ac{CLIP} \ac{VLM} to assign semantic labels to each room instance. Let
$\mathcal{L}^{\text{rcat}}=[\mathbf{l}^{\text{rcat}}_{1},\mathbf{l}^{\text{rcat}}_{2},\dots,\mathbf{l}^{\text{rcat}}_{K}]$ denote the set of predefined room-type categories (e.g., kitchen, bedroom, living room), where $K$ is the number of room types. Using the same \ac{RGB} observations as in the previous stage, the \ac{CLIP} image encoder produces a dense embedding map $\mathcal{F}^{\text{\ac{CLIP}}}\!\in\!\mathbb{R}^{H\times{W}\times{D_{C}}}$, where each pixel $(u, v)$ is represented by a $D_{C}$-dimensional feature vector aligned with natural language concepts. Similarly, the \ac{CLIP} text encoder converts each room category $\mathbf{l}^{\text{rcat}}_{k}$ into a text embedding vector $\mathbf{e}^\text{rcat}_{k}\!\in\!\mathbb{R}^{D_{C}}$, forming the embedding matrix $\mathbf{E}^\text{rcat}\!\in\!\mathbb{R}^{K\times{D_{C}}}$. The embedding map $\mathcal{F}^\text{\ac{CLIP}}$ is flattened into a matrix $\mathbf{Q}^{\text{\ac{CLIP}}}\!\in\!\mathbb{R}^{HW\times{D_{C}}}$. The similarity matrix between each pixel and text embedding is then computed as $\mathbf{S}^\text{\ac{CLIP}}=\mathbf{Q}^{\text{\ac{CLIP}}}(\mathbf{E}^\text{rcat})^{\top}$ where $\mathbf{S}^\text{\ac{CLIP}}\!\in\!\mathbb{R}^{HW\times K}$. The most relevant room label for each pixel $n$ is obtained via

\begin{equation}
\mathbf{r}=\operatorname*{argmax}_{k=1\dots K}\mathbf{S}^{\text{\ac{CLIP}}}_{nk}.
\label{eq:cat_index}
\end{equation}

The resulting vector $\mathbf{r}\!\in\!\mathbb{R}^{HW}$ is reshaped to form an initial room semantic segmentation map $\mathbf{R}\!\in\!\mathbb{R}^{H \times W}$. However, direct pixel-wise classification is sensitive to the agent’s viewpoint, often resulting in noisy predictions that do not consistently reflect the actual room context. To mitigate this issue, the previously generated structural room instance map $\mathcal{S}^\text{final}$ is used to guide the semantic labeling process. For each room segment $\mathbf{s}^\text{final}_b$, the final label for room instance is determined by counting how many times each label appears among all pixels within the segment and selecting the most frequent one:

\begin{figure}[t]
    \centering
    \includegraphics[width=0.55\textwidth]{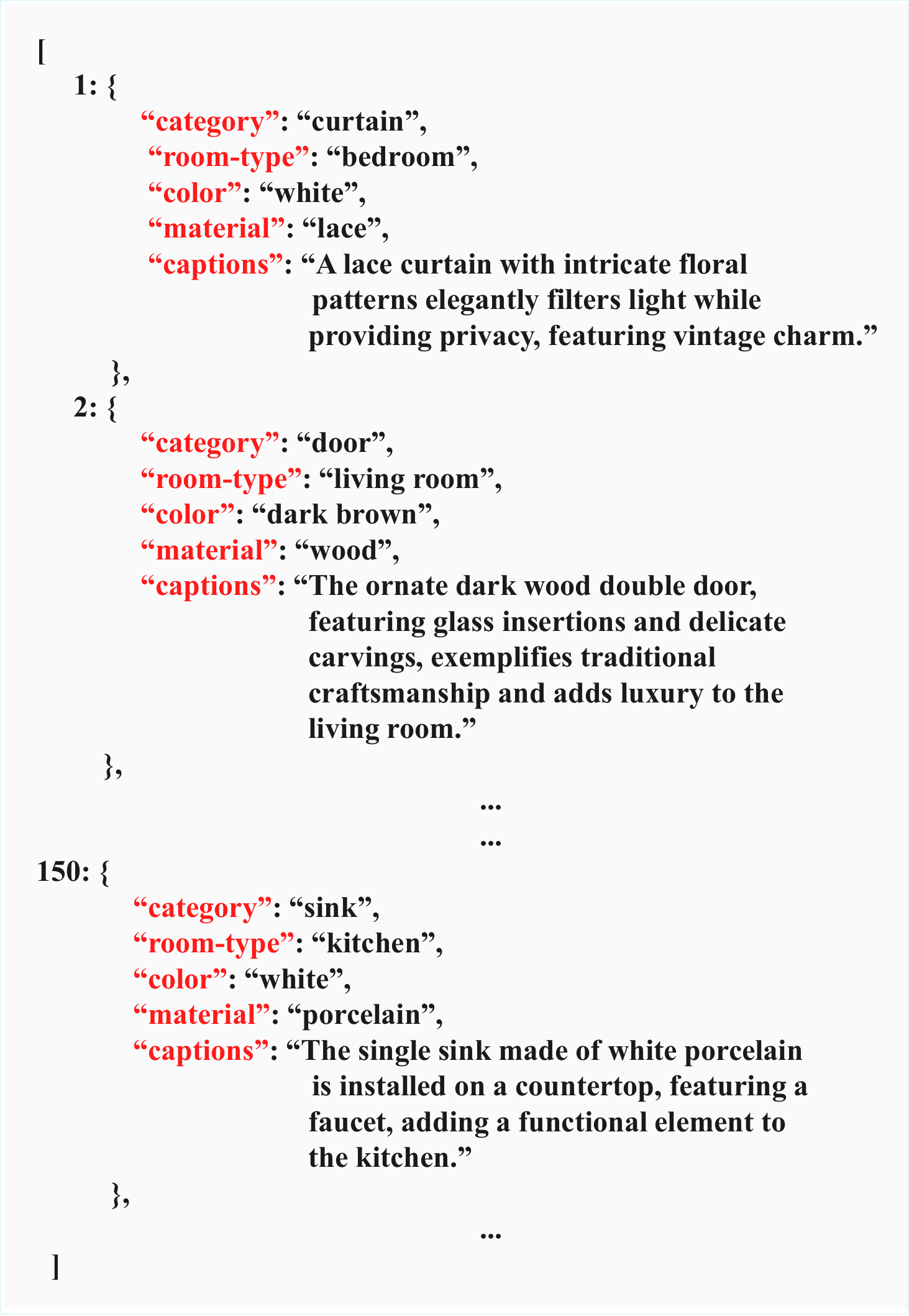}
    \caption{\textbf{Example of instance-level caption annotations. Each detected object instance is associated with semantic attributes including category, room type, color, and material.}}
    \label{fig3}
\end{figure}

\begin{equation}
    l_b = \operatorname*{argmax}_{k}\left| \{ (x, y)\in \mathbf{s}^\text{final}_b \;|\; \mathbf{R}_{xy} = k \} \right|.
\end{equation}

As a result, the final output of the room-level semantic segmentation stage is defined as
\begin{equation}
    \mathcal{R}^\text{final} = \{ (\mathbf{s}^\text{final}_b, l_b)\}_{b=1\dots B},
\end{equation}
where each room segment $\mathbf{s}^\text{final}_b$ is assigned the most frequent label $l_b$ aggregated over its pixels. This region-constrained labeling strategy alleviates pixel-level noise and improves the overall reliability of room-type classification.

\subsection{Instance Captioning}
\label{subsec:methodC}
Detailed textual descriptions are generated for each object instance to enrich the semantic map and enable robust responses to open-ended user queries. The captioning pipeline begins after the instance and room segmentation process. For every detected instance $n$, a set of representative frames $\mathbf{g}^{\text{final}}_{n}$ is selected during the instance segmentation stage. The selection process takes into account both the confidence of the predicted masks and the frequency of the instance’s observations across frames. The selected keyframes are then coupled with the corresponding candidate category $\mathbf{l}^{\text{final}}_n$ obtained from the same stage. Each frame-category pair serves as an input to the \ac{LLaMA} 3.2 model \citep{2024llama3}, which is tasked with producing natural language descriptions. The vision \ac{LLM} is prompted to identify observable object properties such as color, material, texture, and geometric shape. For spatial grounding, each instance is linked to its associated room category. The spatial location of each object is identified on the 2.5D semantic map, and the instance is labeled with the corresponding room label based on the room-level segmentation map $\mathcal{R}^{\text{final}}$. All instance-level captions are then aggregated through \ac{GPT-4} \citep{2023gpt}, which refines them by eliminating redundancy and distilling the essential visual attributes into a compact description. In addition, structured tags are extracted for key attributes.

As illustrated in Fig. \ref{fig3}, the final representation of each instance comprises its room tag providing its spatial context, a natural language caption, and structured key-value tags for efficient indexing and retrieval. This comprehensive representation enables the system to interpret varied open-vocabulary navigation commands, such as "find the small red cup in the kitchen" or "go to the wooden chair in the living room". A detailed description of the prompts and captioning procedure can be found in \ref{sec:appendixA}.

\subsection{Multi-Type Expert Fusion Retrieval for Open-Vocabulary Navigation}
\label{subsec:methodD}
Goal-directed navigation requires identifying the target instance that best satisfies the user's natural language query. However, natural language queries vary significantly in their informational structure, making it challenging to handle them uniformly with a single inference step. We propose the \ac{MTEFR} module, which first constructs an obstacle-aware navigable map and then performs query-adaptive instance selection by routing inference across type-specialized experts and fusing their outputs at the score level.

\subsubsection{Obstacle Map Construction}
The binary obstacle map is primarily constructed using a height threshold. To account for cases where objects should be treated as obstacles even if their mean height falls outside the specified range (e.g., below $h_{\min}$), an open-vocabulary list of potential obstacle categories $\mathcal{L}^{\text{ocat}}=[\mathbf{l}^{\text{ocat}}_{1},\dots,\mathbf{l}^{\text{ocat}}_{K}]$ is derived from natural-language specifications. Using the 2.5D map $\mathcal{M}$ and the instance dictionary $\mathcal{I}^{\text{final}}$, each instance with label $\mathbf{l}^{\text{final}}_{n}$ is treated as an obstacle if either its mean height lies within $[h_{\min},h_{\max}]$ or its label belongs to $\mathcal{L}^{\text{ocat}}$. The binary map $\mathcal{O}\in\{0,1\}^{H\times W}$ is obtained by marking all grid cells that contain at least one such instance.

\subsubsection{Multi-Type Expert Fusion-based Goal Instance Retrieval}
Natural language queries for object-goal navigation can be characterized by the types of information they convey. We model query information along four dimensions: object category (obj), room context (room), descriptive attributes such as color and material generated from instance captions (desc), and abstract semantic intent that omits explicit category names (abs). Queries vary in which of these dimensions are present, ranging from those that rely solely on object labels to those that combine room and descriptive cues without naming the category explicitly. Providing all instance information uniformly to a single LLM conflates these structural differences, distributing attention across irrelevant dimensions and leading to inconsistent performance across query formulations. To address this, the proposed \ac{MTEFR} module dynamically adapts inference to the dimensional structure of each query through a gate LLM and four type-specialized expert \ac{LLMs}, whose outputs are integrated via score-level fusion.

The gate LLM analyzes the input query $\mathcal{Y}$ and outputs a weight vector over four expert types:
\begin{equation}
    \{\omega_e\}_{e \in \mathcal{E}} = \mathrm{LLM}_{\mathrm{gate}}(\mathcal{Y}), 
    \quad 
    \mathcal{E} = \{\mathrm{obj}, \mathrm{room}, \mathrm{desc}, \mathrm{abs}\}
\end{equation}
where $\omega_e \geq 0$ denotes the relative importance of expert type $e$, with $\sum_{e\in\mathcal{E}}\omega_e=1$. For instance, a query emphasizing spatial context and descriptive attributes yields high $\omega_{\text{desc}}$ and $\omega_{\text{room}}$, while a category-explicit query yields high $\omega_{\text{obj}}$. The candidate set $\mathcal{C}$ is constructed from all instances in $\mathcal{I}^{\text{final}}$ and limited to at most $G$ instances to ensure tractable prompt length.

Four expert LLMs then independently rank the candidates in $\mathcal{C}$, each focusing on a distinct information type: the obj expert on category and label matching, the room expert on spatial context, the desc expert on descriptive attributes derived from instance captions, such as color and material, and the abs expert on implicit semantic intent. Each expert returns a top-$\kappa$ ranking, from which a rank score is derived as:
\begin{equation}
    \nu_{e,d} =
    \begin{cases}
        \kappa - \rho_{e,d} + 1 
        & \text{if } d \in \mathrm{top}\text{-}\kappa \text{ of expert } e \\
        0 
        & \text{otherwise}
    \end{cases} \quad,
\end{equation}
where $\rho_{e,d}$ denotes the rank of candidate $d$ assigned by expert $e$.

To balance strong single-expert evidence with multi-expert consensus, two complementary scores are defined and fused at the score level. The weighted sum score aggregates evidence across all expert types, while the peak score captures the strongest evidence from any single expert type:
\begin{equation}
    S^{\mathrm{sum}}_d = \sum_{e \in \mathcal{E}} \omega_e \cdot \nu_{e,d}, 
    \quad 
    P_d = \max_{e \in \mathcal{E}} \left( \omega_e \cdot \nu_{e,d} \right) \quad .
\end{equation}
The final score is then computed as a convex combination of the two, governed by a $\lambda$ schedule:
\begin{equation}
    \lambda = \gamma + (1 - \gamma)\frac{\kappa - 1}{(\kappa - 1) + \mu}\quad,
    \quad
    U_d = \lambda P_d + (1 - \lambda) S^{\mathrm{sum}}_d\quad, 
    \quad 
    d^* = \arg\max_{d \in \mathcal{C}} U_d \quad .
\end{equation}
Here, $\gamma$ controls the base reliance on peak evidence, and $\mu$  governs how rapidly the contribution of the peak score grows with increasing $\kappa$, thereby placing greater emphasis on the strongest single-expert signal as more candidates are considered.

\section{Experiments}
\label{sec:exp}

\label{sec:guidelines}
We evaluate the proposed framework from multiple perspectives. First, we measure instance-level 2.5D mapping accuracy, including a dedicated evaluation of small object preservation, and verify the benefits of the 2.5D representation over 2D mapping (Sec. \ref{subsec:exp1}). Next, we assess room-level segmentation quality (Sec. \ref{subsec:exp2}). We then evaluate goal retrieval and navigation success across diverse query types and conduct ablation studies to quantify the contribution of each module including MTEFR (Sec. \ref{subsec:exp3}). Finally, we compare storage costs across representations (Sec. \ref{subsec:exp4}).

All experiments are conducted within the Habitat simulator \citep{2019habitat}, using \ac{MP3D} \citep{2017mp}, Replica \citep{2019replica}, and \ac{HM3DSEM} \citep{2023hm3dsem} datasets. For fairness, each experiment primarily follows the dataset settings used by the corresponding baselines. The key hyperparameters used throughout all experiments are as follows: $\delta_{\text{geo}} = 0.4, \delta_{\text{sem}} = 0.85, \omega_{\text{geo}} = 0.7, \omega_{\text{sem}} = 0.3, \kappa = 2, \gamma = 0.75, \mu = 16$. These values are fixed across all datasets. The complete hyperparameter list and the prompt templates for the gate LLM and expert LLMs used in MTEFR are available on our project page, which is available at  \url{https://rcilab.github.io/iesm_vln}.



\subsection{2.5D Open-Vocabulary Mapping}
\label{subsec:exp1}

\subsubsection{Instance-Level 2.5D Open-Vocabulary Mapping}
\label{subsubsec:exp1_1}
The goal is to verify whether the mapping process accurately performs object-level segmentation and mapping without over-segmentation or under-segmentation, and whether open-vocabulary features are appropriately stored at the object level.

\noindent \textbf{Baselines and Dataset}: We compare against HOV-SG \citep{2024hovsg} and VLMaps \citep{2023vlmaps}. For HOV-SG (3D instance-wise features), each object point cloud is projected to a top-down mask for a 2.5D comparison. For VLMaps (2D point-wise features), per-instance descriptors are obtained by masking the feature map with ground-truth instance masks and averaging. Following prior setups, HOV-SG uses \ac{CLIP} (ViT-H-14), VLMaps uses LSeg (ViT-B-32), and our method uses \ac{SEEM} (Focal-L). To ensure a fair comparison, we evaluate on the same scenes as the baselines: eight Replica scenes. We additionally evaluate on five \ac{HM3DSEM} scenes to further assess generalizability across diverse environments.

\begin{table}[t]
\scriptsize
\centering
\caption{Instance-level semantic labeling performance under different normalization settings. Instance Difference denotes the absolute difference between ground-truth and predicted object counts. Only instances with an \ac{IoU} greater than 50\% are considered as matched pairs.}
\label{tab1}
\resizebox{\linewidth}{!}{%
\begin{tabular}{|l|l l | c c c c | c c|}
\hline
Dataset & Normalization & Method (Dim.) & $\text{top}_{1}$ & $\text{top}_{5}$ & $\text{top}_{50}$ & $\text{top}_{\text{end}}$ & $\mathbf{AUC}^{\text{top}}$ $\uparrow$ \!\!\!\!\!\!& \textbf{Instance Diff} $\downarrow$ \\
\hline
\multirow{7}{*}{Replica}
&\multirow{2}{*}{Pred-norm.}
 & HOV-SG (3D) & 0.001 & 0.006 & 0.098 & 0.169 & 0.093 & 86.625 \\ 
 & & \textbf{Ours (2.5D)}  & \textbf{0.172} & \textbf{0.295} & \textbf{0.371} & \textbf{0.386} & \textbf{0.356} & \textbf{2.625}\\
\cline{2-9}
& \multirow{2}{*}{GT-norm.}
 & HOV-SG (3D) & 0.003 & 0.016 & 0.244 & \textbf{0.411} & 0.228 & --\\
 & & \textbf{Ours (2.5D)}  & \textbf{0.163} & \textbf{0.280} & \textbf{0.348} & 0.362 & \textbf{0.334} & --\\
\cline{2-9}
& \multirow{3}{*}{Matching-norm.}
& VLMaps + GT mask (2D) & 0.005 & 0.041 & 0.593 & \textbf{1.000} & 0.555 & -- \\
& & HOV-SG (3D) & 0.413 & 0.591 & 0.819 & \textbf{1.000} & 0.817 & --\\
& & \textbf{Ours (2.5D)}   & \textbf{0.447} & \textbf{0.766} & \textbf{0.958} & \textbf{1.000} & \textbf{0.921} & --\\
\hline
\hline
\multirow{7}{*}{HM3DSEM}
&\multirow{2}{*}{Pred-norm.}
 & HOV-SG (3D) & 0.000 & 0.000 & 0.006 & 0.152 & 0.085 & 417.600 \\
 & & \textbf{Ours (2.5D)}  & \textbf{0.081} & \textbf{0.179} & \textbf{0.291} & \textbf{0.396} & \textbf{0.371} & \textbf{67.4}\\
\cline{2-9}
& \multirow{2}{*}{GT-norm.}
 & HOV-SG (3D) & 0.000 & 0.000 & 0.018 & \textbf{0.420} & 0.238 & --\\
 & & \textbf{Ours (2.5D)} & \textbf{0.057}  & \textbf{0.124} & \textbf{0.201} & 0.271 & \textbf{0.255} & --\\
\cline{2-9}
& \multirow{3}{*}{Matching-norm.}
& VLMaps + GT mask (2D) & 0.000 & 0.000 & 0.043 & \textbf{1.000} & 0.564 & -- \\
& & HOV-SG (3D) & 0.171 & 0.271 & 0.655 & 1.000 & 0.914 & --\\
& & \textbf{Ours (2.5D)}  & \textbf{0.210} & \textbf{0.459} & \textbf{0.742} & \textbf{1.000} & \textbf{0.941} & --\\
\hline
\end{tabular}
}
\end{table}

\begin{figure}[t]
    \centering
    \includegraphics[width=1.0\textwidth]{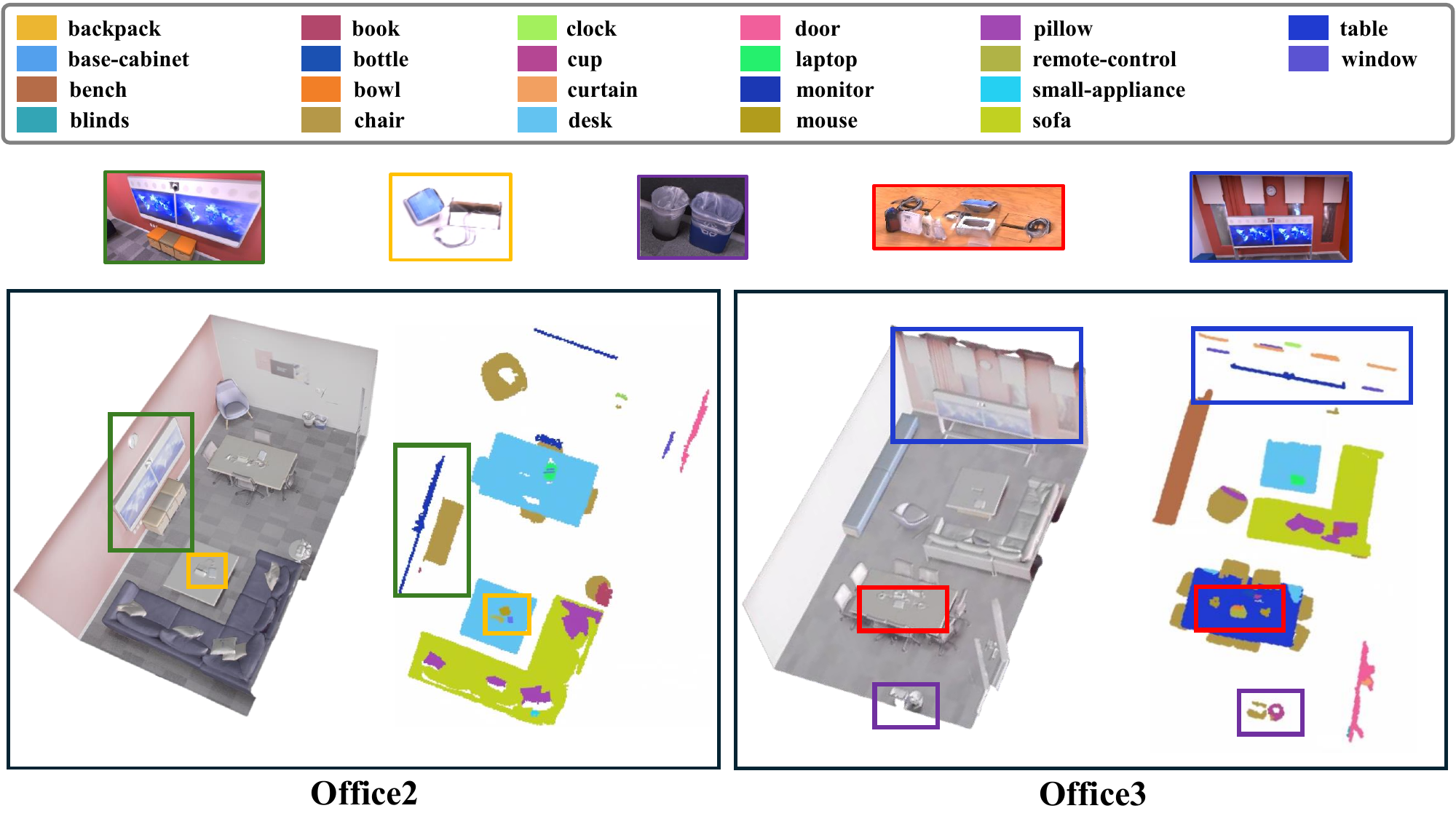}
    \caption{Instance-level 2.5D open-vocabulary mapping results on two scenes from the Replica dataset. For each scene, the left side shows a 3D rendered view of the environment and the right side shows the corresponding top-down 2.5D semantic map, where each instance is color-coded according to the category legend at the top. The colored bounding boxes highlight small objects (e.g., bottles, laptops, cups) that are reliably preserved and accurately mapped without disappearing or being absorbed into larger neighboring instances.}
    \label{fig4}
\end{figure}

\noindent \textbf{Metrics}: Following HOV-SG, we adopt the labeling-based $\text{AUC}^{\text{top}}$ for semantic evaluation at the object level. This metric computes the cumulative ratio of correctly labeled instance matches within the top-$k$ predictions, defined as
\begin{equation}
f(k) = \frac{|S_k|}{|M|}, \quad 
\text{AUC}^{\text{top}} = \int_0^1 f(x)\, dx,
\end{equation}
where $|S_k|$ is the number of pairs correctly labeled among the top-$k$, and $|M|$ is the total number of matched pairs. Here, the horizontal axis is rescaled to $[0,1]$ by the total number of categories, so that $k$ is mapped to the unit interval, enabling the computation of the \ac{AUC}. Since $f(k)$ is discrete, we interpolate it over the unit interval to obtain a continuous curve for AUC computation, and the same interpolation procedure is applied consistently to all AUC variants introduced below. 

However, instance-level evaluation inherently involves two stages: instance matching and label assignment. The original HOV-SG formulation normalizes by the number of matched pairs $|M|$, which does not explicitly penalize redundant predictions or over-segmentation errors, since the denominator is restricted to the matched subset rather than the entire prediction set.

\begin{table}[t]
\scriptsize
\centering
\caption{Small object instance-level semantic labeling performance on the Replica and 
\ac{HM3DSEM} datasets. Small objects are defined as instances belonging to categories whose top-down projected occupancy area falls within the bottom 20\% across all instances in the dataset. Only predicted and ground-truth instances with an \ac{IoU} greater than 50\% are considered as matched pairs.
}
\label{tab:small}
\resizebox{\linewidth}{!}{%
\begin{tabular}{|l|l l | c c c c | c c|}
\hline
Dataset & Normalization & Method (Dim.) & $\text{top}_{1}$ & $\text{top}_{5}$ & $\text{top}_{50}$ & $\text{top}_{\text{end}}$ & $\mathbf{AUC}^{\text{top}}$ $\uparrow$ \!\!\!\!\!\!& \textbf{Instance Diff} $\downarrow$ \\
\hline
\multirow{7}{*}{Replica}
&\multirow{2}{*}{Pred-norm.}
 & HOV-SG (3D) & 0.001 & 0.003 & 0.069 & 0.098 & 0.060 & 100.875 \\
 & & \textbf{Ours (2.5D)}  & \textbf{0.107} & \textbf{0.158} & \textbf{0.199} & \textbf{0.202} & \textbf{0.189} & \textbf{11.625}\\
\cline{2-9}
& \multirow{2}{*}{GT-norm.}
 & HOV-SG (3D) & 0.003 & 0.017 & 0.235 & \textbf{0.346} & 0.208 & --\\
 & & \textbf{Ours (2.5D)}  & \textbf{0.152} & \textbf{0.222} & \textbf{0.278} & 0.283 & \textbf{0.265} & --\\
\cline{2-9}
& \multirow{3}{*}{Matching-norm.}
& VLMaps + GT mask (2D) & 0.007 & 0.055 & 0.694 & \textbf{1.000} & 0.611 & -- \\
& & HOV-SG (3D) & 0.413 & 0.591 & 0.830 & \textbf{1.000} & 0.813 & --\\
& & 
\textbf{Ours (2.5D)}   & \textbf{0.534} & \textbf{0.775} & \textbf{0.982} & \textbf{1.000} & \textbf{0.933} & --\\
\hline
\hline
\multirow{7}{*}{HM3DSEM}
&\multirow{2}{*}{Pred-norm.}
 & HOV-SG (3D) & 0.000 & 0.000 & 0.001 & 0.100 & 0.050 & 477.000 \\
 & & \textbf{Ours (2.5D)}  & \textbf{0.036} & \textbf{0.094} & \textbf{0.138} & \textbf{0.200} & \textbf{0.186} & \textbf{8.000}\\
\cline{2-9}
& \multirow{2}{*}{GT-norm.}
 & HOV-SG (3D) & 0.000 & 0.000 & 0.003 & \textbf{0.373} & \textbf{0.192} & --\\
 & & \textbf{Ours (2.5D)}  & \textbf{0.034} & \textbf{0.088} & \textbf{0.129} & 0.187 & 0.174 & --\\
\cline{2-9}
& \multirow{3}{*}{Matching-norm.}
& VLMaps + GT mask (2D) & 0.000 & 0.000 & 0.009 & \textbf{1.000} & 0.510 & -- \\
& & HOV-SG (3D) & 0.149 & 0.261 & \textbf{0.707} & \textbf{1.000} & 0.932 & --\\
& & 
\textbf{Ours (2.5D)}   & \textbf{0.180} & \textbf{0.468} & 0.695 & \textbf{1.000} & \textbf{0.933} & --\\
\hline
\end{tabular}
}
\end{table}

To overcome this limitation, we introduce additional denominator choices for normalization. Here, normalized refers to the choice of denominator. Prediction-normalized labeling \ac{AUC} ($\text{AUC}_{\text{pred}}^{\text{top}}$) penalizes redundant or spurious predictions by normalizing with respect to the total number of predicted objects:
\begin{equation}
f_{\text{pred}}(k) = \frac{|S_k|}{|P|}, \quad 
\text{AUC}_{\text{pred}}^{\text{top}} = \int_0^1 f_{\text{pred}}(x)\, dx,
\end{equation}
where $|P|$ denotes the total number of predictions.
Ground-truth-normalized labeling AUC ($\text{AUC}_{\text{gt}}^{\text{top}}$) penalizes under-segmentation and missed detections by normalizing with respect to the total number of ground-truth objects:   
\begin{equation}
f_{\text{gt}}(k) = \frac{|S_k|}{|G|}, \quad 
\text{AUC}_{\text{gt}}^{\text{top}} = \int_0^1 f_{\text{gt}}(x)\, dx,
\end{equation}
where $|G|$ denotes the total number of ground-truth objects.
Matching-normalized labeling \ac{AUC} ($\text{AUC}_{\text{match}}^{\text{top}}$) evaluates whether object-level semantics are preserved without additional penalties by normalizing with respect to the number of matched pairs:
\begin{equation}
f_{\text{match}}(k) = \frac{|S_k|}{|M|}, \quad 
\text{AUC}_{\text{match}}^{\text{top}} = \int_0^1 f_{\text{match}}(x)\, dx.
\end{equation}

\noindent \textbf{Results}: Fig. \ref{fig4} presents qualitative results of our 2.5D open-vocabulary mapping on the Replica dataset, confirming that objects in the environment, including small ones, are accurately mapped without being overlooked. Table \ref{tab1} reports values at different top-$k$ thresholds, the overall $\text{AUC}^{\text{top}}$ score, and the average absolute difference between predicted and ground-truth instance counts. Since VLMaps is point-wise, masking its feature map with ground-truth instance masks produces a predicted instance count that matches the ground truth by construction. Therefore, only the matching-normalized metric is reported for VLMaps.

\begin{table}[h]
\centering
\caption{Evaluation of semantic mapping performance against 2D-based models on the MP3D and Replica datasets.}
\label{tab2}

\resizebox{\linewidth}{!}{%
\begin{tabular}{|l l|c c c c|c c c c|}
\hline
\multirow{2}{*}{\textbf{Method}} & 
\multirow{2}{*}{\textbf{VLM (Backbone)}} & 
\multicolumn{4}{c|}{\textbf{MP3D}} & 
\multicolumn{4}{c|}{\textbf{Replica}} \\
\cline{3-10}
 & & Acc & mAcc & mIoU & F-mIoU & Acc & mAcc & mIoU & F-mIoU \\
\hline
\multirow{1}{*}{\ac{CLIP} Map} 
 & \ac{CLIP} (ViT-B-32)  & 0.106 & 0.102 & 0.017 & 0.026 & 0.147 & 0.066 & 0.012 & 0.028 \\
\hline
\multirow{2}{*}{VLMaps} 
 & LSeg (ViT-B-32) & 0.448 & 0.304 & 0.198 & 0.330 & 0.485 & 0.260 & 0.175 & 0.359 \\
 & \ac{SEEM} (Focal-L)  & 0.447 & 0.310 & 0.226 & 0.367 & \textbf{0.503} & 0.272 & 0.200 & 0.357 \\
\hline
\multirow{1}{*}{Ours (w/o secondary fusion)} 
 & \ac{SEEM} (Focal-L) & 0.434     & 0.306     & 0.215     & 0.344     & 0.448     & 0.261     & 0.179     & 0.337     \\
\hline
\multirow{1}{*}{\textbf{Ours (w/ secondary fusion)}} 
 & \ac{SEEM} (Focal-L) & \textbf{0.507} & \textbf{0.360} & \textbf{0.238} & \textbf{0.394} & 
 0.489 & \textbf{0.284} & \textbf{0.202} & \textbf{0.380} \\
\hline
\end{tabular}
}
\end{table}

\begin{figure}[t]
\centering
\subfloat[2D mapping results on scene 2t7WUuJeko7 of the MP3D.\label{fig5_a}]{
    \includegraphics[width=0.92\textwidth]{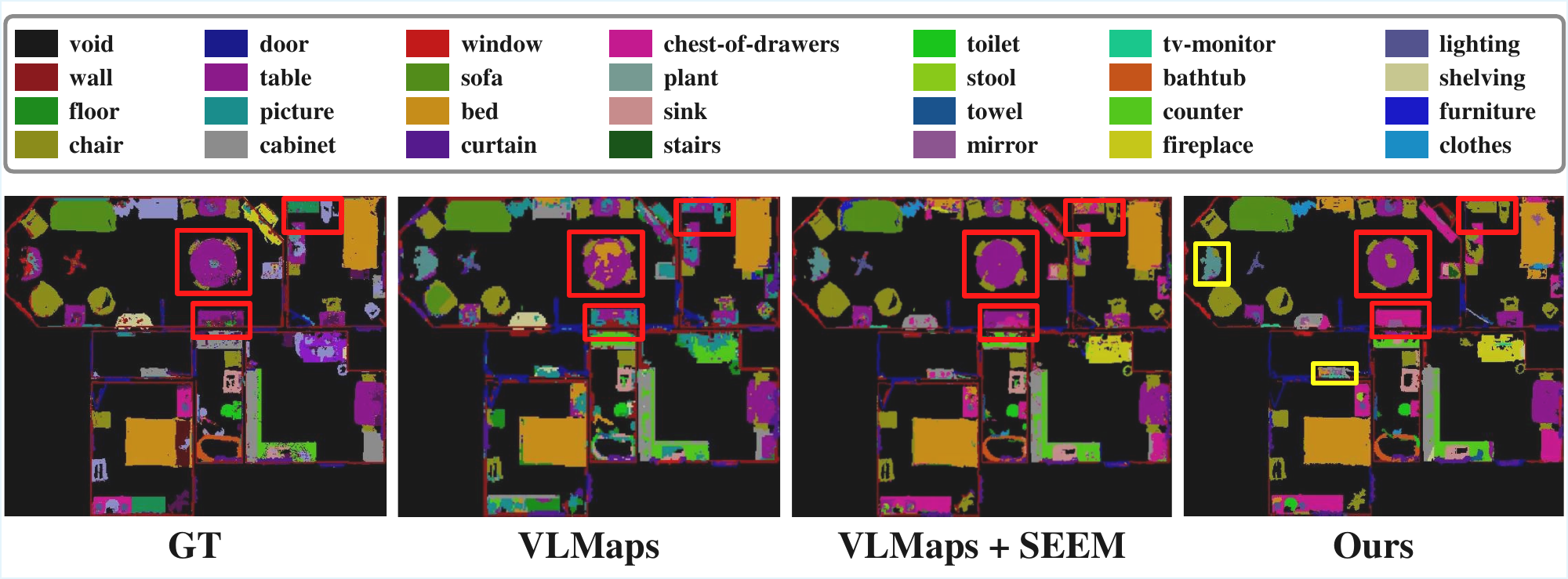}
}
\vspace{0.1em}
\subfloat[2D mapping results on scene apartment\_0 of the Replica.\label{fig5_b}]{
    \includegraphics[width=0.92\textwidth]{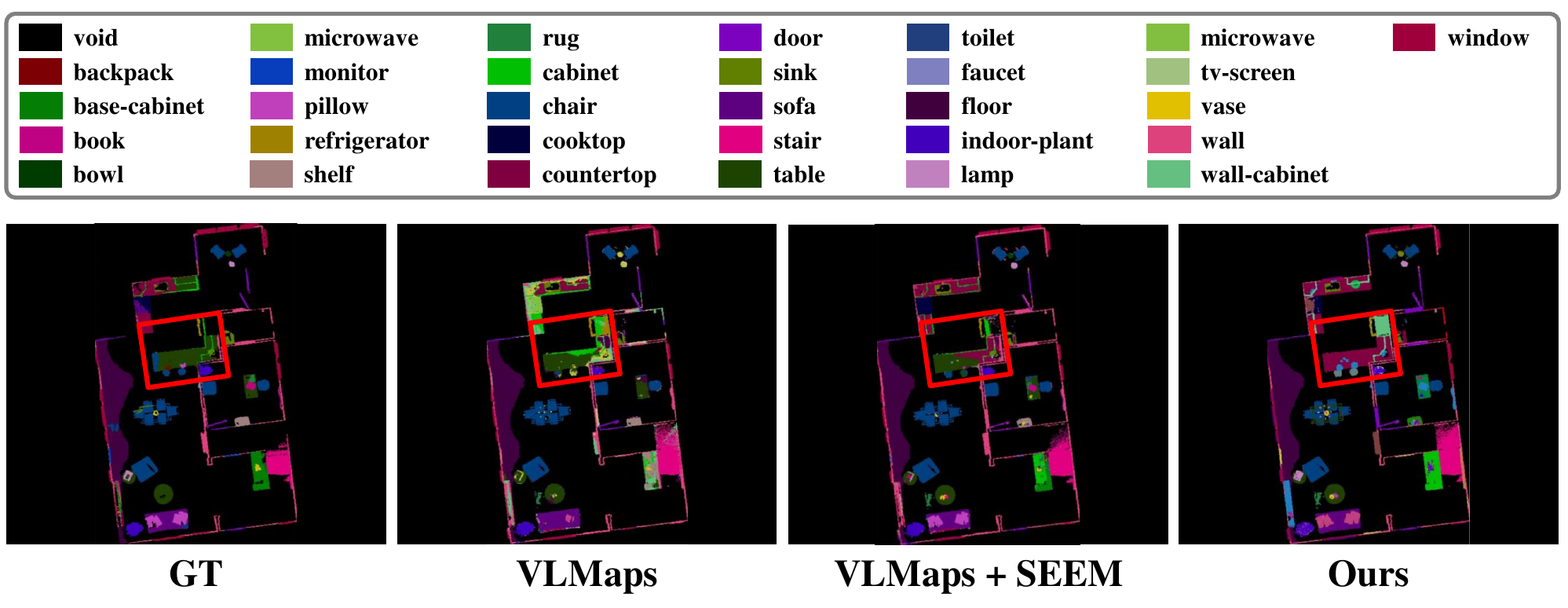}
}
\caption{2D mapping results of each method. The red boxes highlight that our method preserves objects as coherent instances without fragmentation. This demonstrates that the 2.5D mapping, when projected into a top-down view, mitigates the semantic ambiguity caused by vertically overlapping object features in conventional 2D projections. The yellow boxes indicate cases where a single object is split into multiple fragments due to insufficient IoU between the object captured in the affected frame and the corresponding mapped instance, caused by abrupt viewpoint changes during data acquisition. While such over-segmentation artifacts still occasionally occur in our method, their frequency is substantially lower compared to the baselines.}
\label{fig5}
\end{figure}

For the $\text{AUC}_{\text{pred}}^{\text{top}}$, our method exceeds HOV-SG across both datasets, achieving an AUC of 0.356 versus 0.093 for HOV-SG on Replica and 0.371 versus 0.085 on 
\ac{HM3DSEM}. This gap highlights our robustness to spurious predicted instances caused by over-segmentation and redundant small-object predictions. HOV-SG suffers from low robustness due to SAM+\ac{CLIP}'s over-segmentation issues and poor integration of small-sized objects. In contrast, our method avoids over-segmentation through \ac{SEEM}'s panoptic segmentation model and maintains better preservation of small-sized objects. Consistently, the per-scene mean absolute instance-count gap is 2.6 for our method versus 86.6 for HOV-SG on Replica, and 67.4 versus 417.6 on \ac{HM3DSEM}.

For the $\text{AUC}_{\text{gt}}^{\text{top}}$, our method attains higher overall AUC scores on both datasets, indicating comparable or better recall of ground-truth instances across the full ranking range.

For the $\text{AUC}_{\text{match}}^{\text{top}}$, our method achieves 0.921 on Replica and 0.941 on 
\ac{HM3DSEM}, outperforming HOV-SG (0.817 and 0.914, respectively). VLMaps performs markedly worse than both HOV-SG and our method even with ground-truth masking (0.555 on Replica and 0.564 on \ac{HM3DSEM}), due to the 2D point-wise storage of features where instance representations within a grid cell are blended under vertical overlap. Averaged across both datasets and all applicable normalization schemes, our method consistently outperforms the baselines, confirming the generalizability of the proposed 2.5D representation across diverse environments.

\subsubsection{Small Object Preservation}
Small objects are particularly challenging for mapping systems, as they are prone to being absorbed into neighboring instances or missed entirely during the top-down projection process. We evaluate whether the proposed 2.5D mapping reliably preserves such objects compared to the baselines.

\noindent \textbf{Baselines and Dataset}: The same baselines, datasets, and evaluation protocol as in Sec. \ref{subsubsec:exp1_1} are used. Small objects are defined per dataset by examining the ground-truth instance annotations across all scenes and identifying categories whose top-down projected occupancy area falls within the bottom 20\% across all instances in the dataset. Only instances belonging to these categories are included in the evaluation.

\noindent \textbf{Metrics}: The same three normalization variants of $\text{AUC}^{\text{top}}$ and Instance Diff as defined in Sec. \ref{subsubsec:exp1_1} are used. 

\noindent \textbf{Results}: Table \ref{tab:small} reports the small object mapping performance.

For the $\text{AUC}_{\text{pred}}^{\text{top}}$, the proposed method outperforms HOV-SG on both datasets, achieving 0.189 versus 0.060 on Replica and 0.186 versus 0.050 on \ac{HM3DSEM}. The instance-count gap further confirms this trend, with our method showing 11.6 versus 100.9 on Replica and 8.0 versus 477.0 on \ac{HM3DSEM}, demonstrating that our method avoids the over-segmentation and small object absorption issues that affect HOV-SG.

For the $\text{AUC}_{\text{gt}}^{\text{top}}$, our method outperforms HOV-SG on Replica (0.265 vs 0.208). On \ac{HM3DSEM}, however, HOV-SG achieves a slightly higher AUC (0.192 vs 0.174). This is attributed to the larger scene scale of \ac{HM3DSEM}: while our method effectively distinguishes small objects from one another in compact, densely arranged regions, small objects in larger scenes occasionally tend to be absorbed into adjacent larger instances, representing a failure case in such environments. Despite this, the performance gap remains small, and the substantially larger advantage in the prediction-normalized metric confirms that our method maintains superior precision in the instances it does detect.

For the $\text{AUC}_{\text{match}}^{\text{top}}$, the proposed method achieves 0.933 on both Replica and \ac{HM3DSEM}, outperforming HOV-SG (0.813 and 0.932, respectively). VLMaps performs markedly worse on both datasets (0.611 on Replica, 0.510 on \ac{HM3DSEM}), reflecting the fundamental limitation of 2D point-wise storage in preserving small object semantics under vertical overlap. Across both datasets and all normalization schemes, our method consistently outperforms the baselines, confirming the robustness of the 2.5D representation for small object preservation.

\subsubsection{Additional Validation: Effectiveness of 2.5D Instance Mapping over 2D}
In this section, we further investigate whether the 2.5D formulation provides advantages over a 2D mapping approach. To this end, we project our 2.5D map into a top-down 2D representation and compare it with purely 2D grid maps.

\noindent \textbf{Baselines and Dataset}: We compare against representative 2D projection baselines: VLMaps (LSeg, ViT-B-32), VLMaps+\ac{SEEM}, which replaces LSeg with \ac{SEEM} (Focal-L), and \ac{CLIP} Map (\ac{CLIP}, ViT-B-32), which encodes the environment using \ac{CLIP}-derived representations. We also report our framework with and without secondary-instance fusion to verify the necessity of this step. To ensure a fair comparison, experiments are conducted on the same MP3D dataset as used in the baselines, along with the Replica dataset, using four representative scenes per dataset.

\noindent \textbf{Metrics}: Evaluation is conducted in a point-wise manner by comparing the predicted grid map with the ground-truth grid map. Specifically, we report four metrics. \ac{Acc} measures overall accuracy across all grid cells, \ac{mAcc} is the mean class accuracy averaged over categories, \ac{mIoU} represents the mean \ac{IoU} across categories, and \ac{F-mIoU} accounts for category frequency.

\begin{table}[t]
\centering
\caption{Evaluation of the room segmentation on the MP3D dataset.}
\label{tab3}
\resizebox{0.3\columnwidth}{!}{
\begin{tabular}{|l|c|c|c|}
\hline
\multirow{2}{*}{\textbf{Method}} & \multicolumn{3}{c|}{\textbf{MP3D}} \\
\cline{2-4}
 & Acc & mAcc & mIoU\\
\hline
\ac{CLIP} & 0.73 & 0.72 & 0.53 \\
\textbf{Ours} & \textbf{0.89} & \textbf{0.86} & \textbf{0.75} \\
\hline
\end{tabular}
}
\end{table}

\begin{figure}[t]
\centering
\subfloat[Room segmentation result on scene 2t7WUuJeko7 of the MP3D.\label{fig6_a}]
{\includegraphics[width=0.47\textwidth]{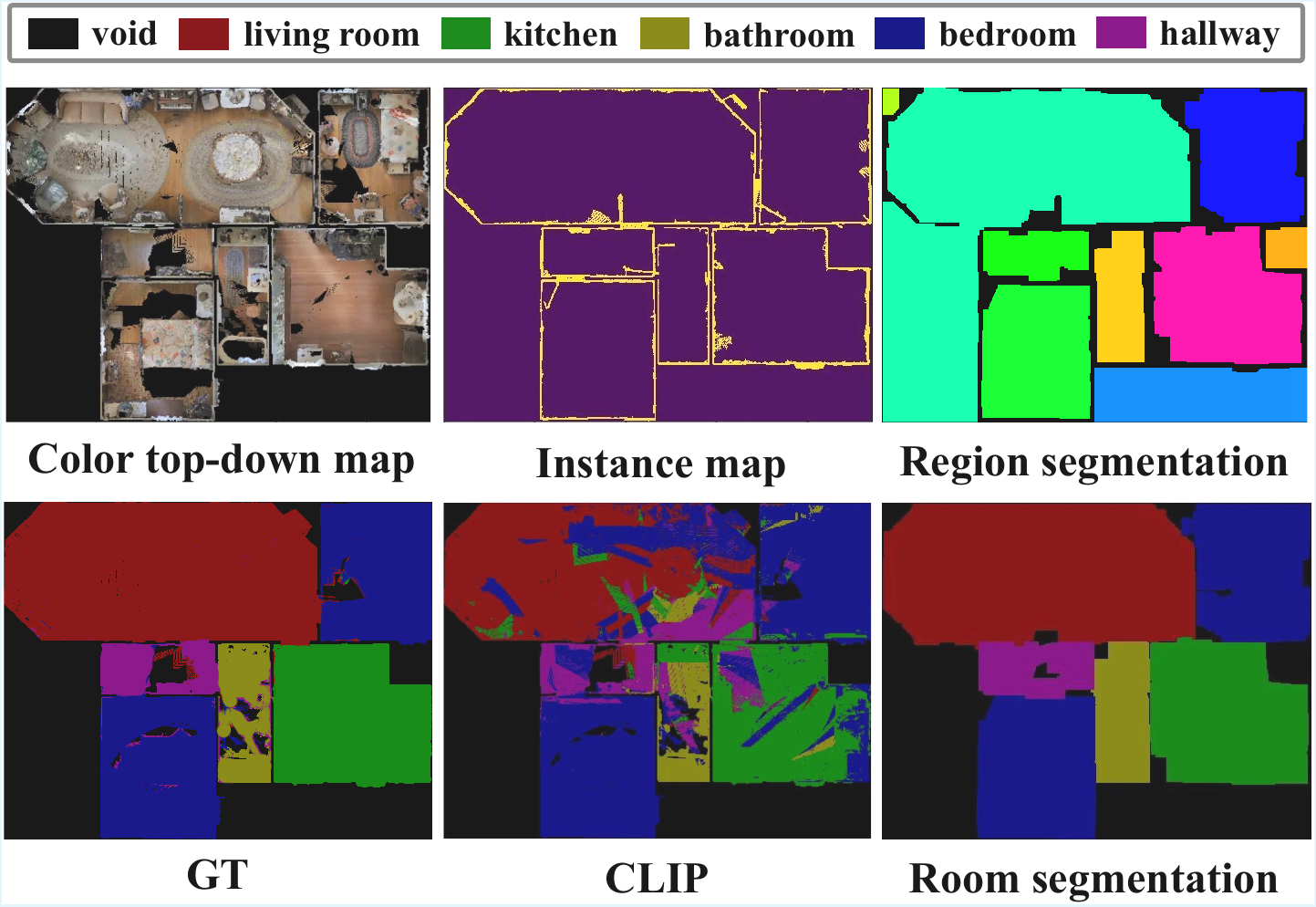}}
\hspace{0.03\textwidth}
\subfloat[Room segmentation result on scene RPmz2sHmrrY of the MP3D.\label{fig6_b}]
{\includegraphics[width=0.47\textwidth]{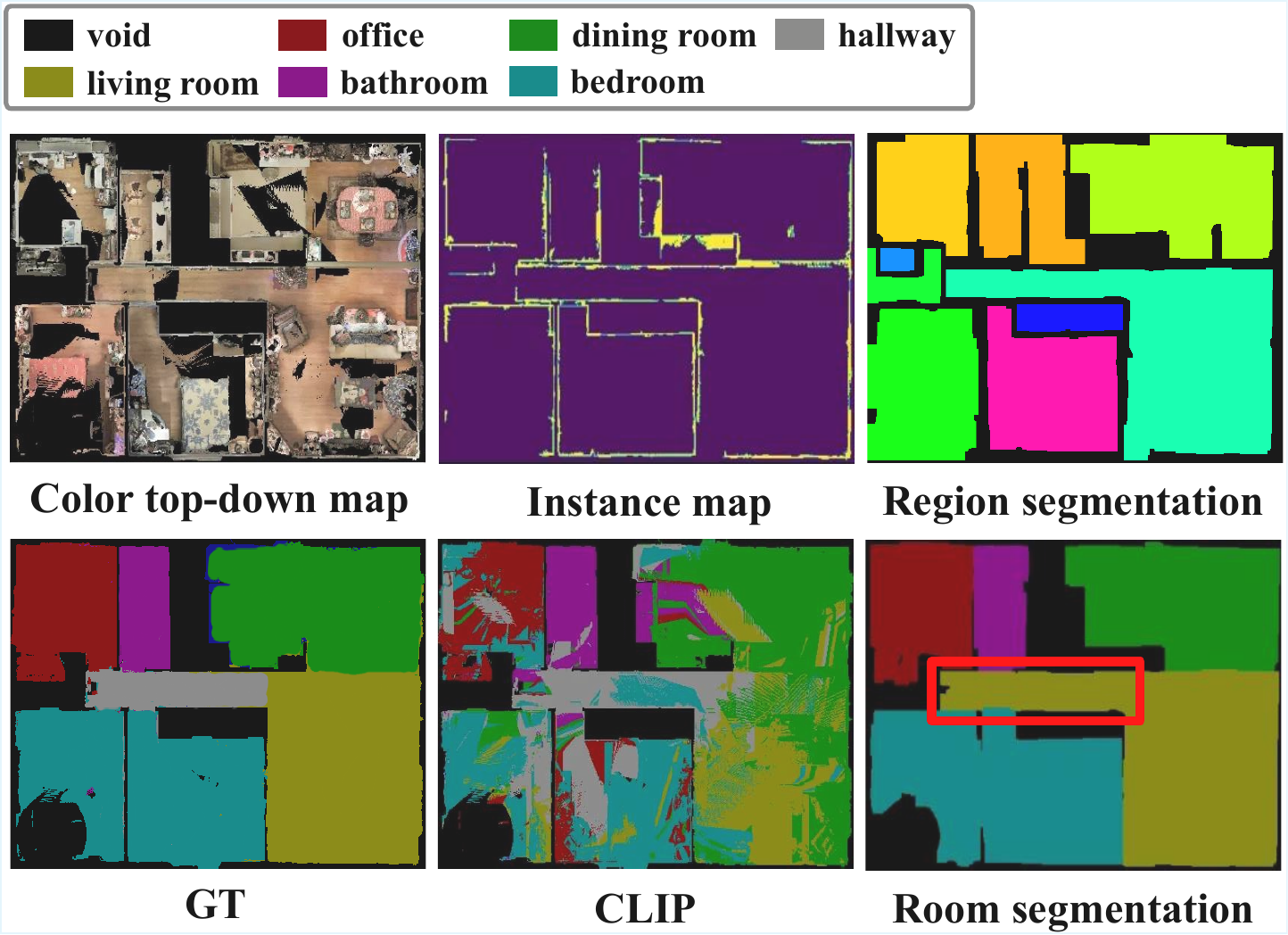}}
\caption{Room segmentation results on two scenes from the MP3D dataset. The top row shows the colorized top-down map, predicted instance map, and region segmentation. The bottom row 
compares the \ac{GT}, \ac{CLIP}-based semantic segmentation, and our final room-level 
segmentation output. The red box highlights a representative failure 
region caused by this structural absence.}
\label{fig6}
\end{figure}

\noindent \textbf{Results}: The quantitative results are summarized in Table \ref{tab2}. Across most metrics, the performance follows the order: Ours, VLMaps+\ac{SEEM}, VLMaps, and \ac{CLIP} Map. Unlike models with pixel-level or object-level embeddings, \ac{CLIP} Map uses an image-level embedding. As a result, the features stored in the map are intermixed across pixels and instances, which lowers performance. The comparison between VLMaps and VLMaps+\ac{SEEM} highlights the superior open-vocabulary capability of the \ac{SEEM} visual encoder. Meanwhile, the improvement of our method over VLMaps+\ac{SEEM} demonstrates the effectiveness of 2.5D mapping with instance-level integration. By preserving vertical distinctions that are collapsed in 2D projections, our method reduces semantic ambiguity and prevents erroneous category assignments. The qualitative results in Fig. \ref{fig5} also confirm this trend, showing that our method produces more consistent object-level segmentations compared to the baselines.

In Fig. \ref{fig5_b}, the region marked with a red box is labeled as "table" in the ground truth, while predicted as "countertop". This mismatch reflects the open-vocabulary nature of the task. The object is indeed a kitchen countertop, and such cases demonstrate that a single object may correspond to multiple semantically valid labels depending on the context.

Despite these advantages, our method is not entirely free from segmentation artifacts. As indicated by the yellow boxes in Fig. \ref{fig5}, instances of over-segmentation, where a single object is split into multiple fragments, still occasionally occur in our method as well. While our 2.5D representation with \ac{SEEM}-based panoptic segmentation reduces such cases compared to the baselines, abrupt viewpoint changes during the agent's data acquisition process can cause the object captured by \ac{SEEM} in the affected frame to yield insufficient IoU with the corresponding mapped instance, leading it to be registered as a separate instance rather than merged with the existing one. This represents an inherent challenge tied to the data acquisition process rather than a fundamental limitation of the proposed representation itself.

Overall, the comparison against 2D projection-based approaches confirms that our 2.5D framework not only improves accuracy but also provides more reliable and semantically coherent object-level representations in complex scenes.

\begin{figure}[t]
    \centering
    \includegraphics[width=0.98\textwidth]{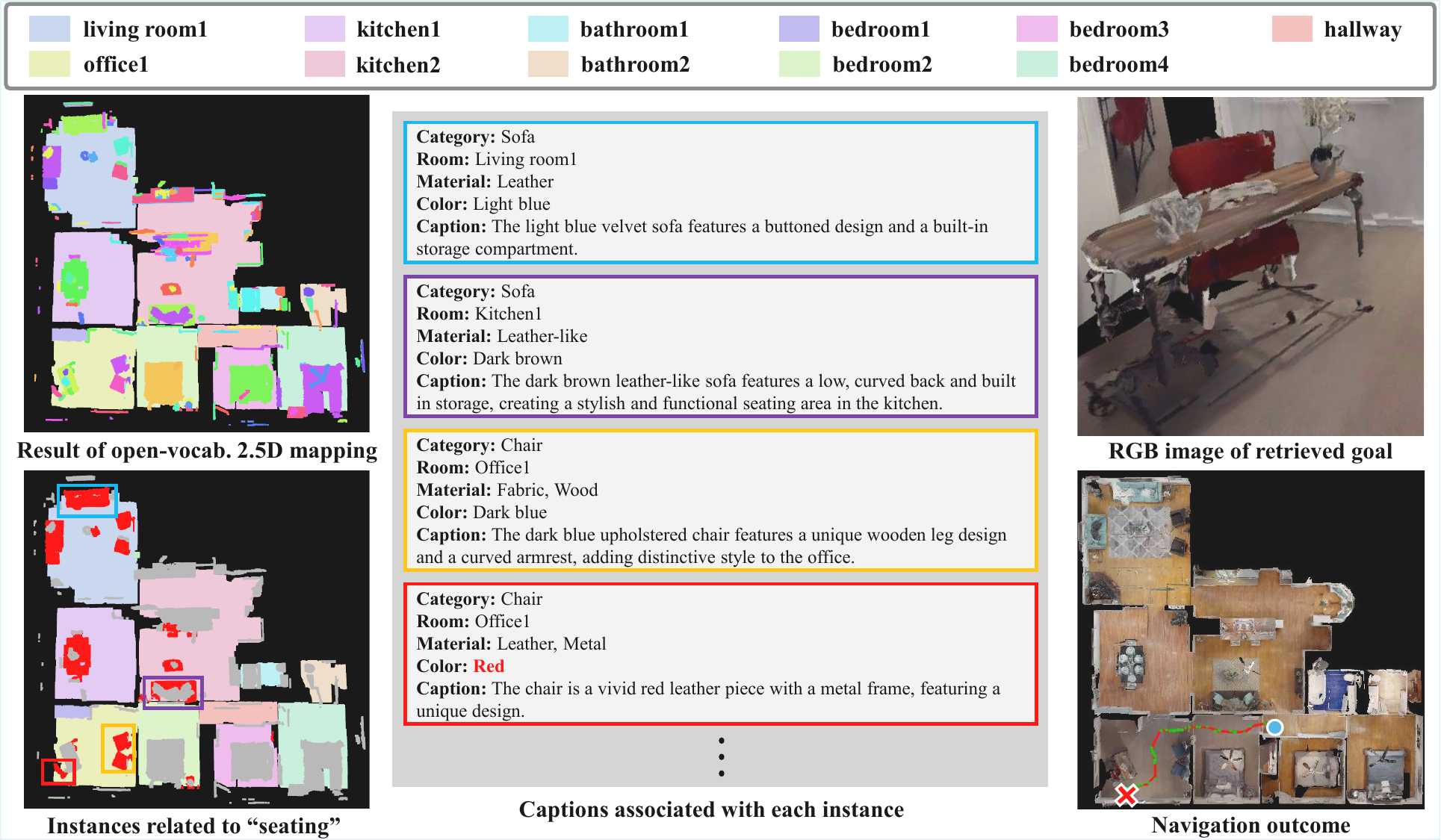}
    \caption{Qualitative navigation result for the abstract query, "Navigate to the red seating object in the office," on scene 00824 of the \ac{HM3DSEM} dataset. \textbf{(Left)} The 2.5D semantic map with room segmentation and instances associated with the abstract concept of "seating." \textbf{(Center)} Full instance-level information stored in the map for each candidate, including category, room type, color, material, and natural language captions. Each expert LLM returns ranked candidates based on its corresponding information type, and score-level fusion determines the final retrieval target. \textbf{(Right)} The retrieved goal instance and navigation outcome, where the red cross marks the goal and the blue circle the starting position. The correct target, a red leather chair in Office1, is successfully retrieved and reached.}
    \label{fig7}
\end{figure}

\subsection{Room Semantic Segmentation}\label{subsec:exp2}
The purpose is to assess whether room segmentation achieves spatially consistent partitions that align with structural cues, and whether the resulting regions retain accurate open-vocabulary semantic labels.

\noindent \textbf{Baselines and Dataset}: As a baseline, \ac{CLIP}-based point features are projected onto the top-down map and used for pixel-wise room-type classification via similarity with predefined text prompts. Our proposed approach is quantitatively evaluated on the MP3D dataset, using two scenes.

\noindent \textbf{Metrics}: The evaluation is conducted using standard metrics including \ac{Acc}, \ac{mAcc}, and \ac{mIoU}.

\noindent \textbf{Results}: Our method incorporates a structural room instance map, derived from architectural cues such as walls and doors, to guide region-wise label assignment based on \ac{CLIP}. This structural constraint significantly reduces pixel-level noise and enhances spatial coherence in room segmentation. As summarized in Table \ref{tab3}, our method consistently outperforms the \ac{CLIP} baseline across all evaluation metrics, improving \ac{Acc} from 0.73 to
0.89, \ac{mAcc} from 0.72 to 0.86, and \ac{mIoU} from 0.53 to 0.75. The largest absolute gain is observed in \ac{Acc} and \ac{mAcc}, reflecting the benefit of spatially contiguous region-constrained labeling over pixel-wise classification. We observe that \ac{CLIP}-based segmentation suffers from semantic fragmentation and spatial inconsistency, often mislabeling adjacent areas due to the agent’s limited viewpoint. In contrast, our method successfully corrects this issue by enforcing spatial contiguity and leveraging structural priors such as walls and doorframes, resulting in well-delineated and semantically coherent segmentation.

Despite these advantages, our method is not entirely free from failure cases. Fig. \ref{fig6} illustrates qualitative segmentation results on two scenes from the MP3D dataset. In Fig. \ref{fig6_a}, physical walls and doorframes provide well-defined boundaries, and the proposed structural prior yields consistent gains across all room categories. In contrast, Fig. \ref{fig6_b} contains an open-plan area where the hallway and living room share no physical partition, which reveals a systematic limitation of the LSD-based boundary detection stage. As highlighted by the red bounding box, the \ac{CLIP}
baseline correctly identifies the hallway as a distinct semantic region at the pixel
level; however, because no physical partition separates the hallway from the neighboring
living room, LSD fails to generate a boundary segment between them. Consequently,
connected component analysis merges the two spaces into a single region, and
majority-vote labeling assigns the living room label to the entire merged area, causing
the hallway to disappear from the final output. This case demonstrates that the failure
originates in the structural separation stage rather than in the semantic recognition
capability of \ac{CLIP}.

Nevertheless, the overall quantitative results confirm that the proposed method achieves consistent improvements over the \ac{CLIP} baseline across all metrics, demonstrating its effectiveness in producing spatially coherent room segmentation.

\subsection{Navigation}\label{subsec:exp3}
We evaluate the end-to-end navigation performance of the proposed framework across diverse natural language query formulations. The evaluation covers both qualitative and quantitative aspects: Sec. \ref{subsubsec:exp3_1} presents a qualitative example illustrating how the framework operates from map construction to goal-directed navigation, and Sec. \ref{subsubsec:exp3_2} provides quantitative comparisons and ablation studies.

\subsubsection{Qualitative Evaluation}\label{subsubsec:exp3_1}
Fig. \ref{fig7} presents a qualitative evaluation on the most challenging query type, (a,r,d), where the target is referred to through an abstract expression without an explicit category name. For the query "Navigate to the red seating object in the office," the center panel shows the full instance-level information stored in the map for each candidate, including category, room type, color, material, and natural language captions. Each expert LLM returns ranked candidates based on its corresponding information type, and their outputs are integrated through score-level fusion to determine the final retrieval target. The left panel visualizes instances associated with the abstract concept of "seating" on the 2.5D semantic map, confirming that seating-related candidates such as chairs and sofas across multiple rooms are successfully identified. As shown in the right panel, the correct target, a red leather chair in Office1, is successfully retrieved and the agent reaches the goal, demonstrating that the framework operates correctly from object retrieval through to navigation. Quantitative evaluation is presented in Sec. \ref{subsubsec:exp3_2}.

\subsubsection{Quantitative Evaluation}\label{subsubsec:exp3_2}
We quantitatively evaluate goal retrieval and navigation success across five query types on five \ac{HM3DSEM} scenes. For each scene and for every query type, 50 trials are conducted by randomly sampling object instances from the 30 most frequent categories, with a starting position randomly assigned per instance, resulting in 250 trials per query type in total across the five scenes. To ensure fair comparison across methods, the same starting position is used for all methods evaluated on the same instance. Queries are automatically generated for each of the five types as described in Sec. \ref{sec:appendixC}.

\noindent \textbf{Baselines and Dataset}: We compare VLMaps, HOV-SG, and our proposed method using the 
\ac{HM3DSEM} dataset. For multi-floor scenes, we evaluate only the first floor, as both VLMaps and our method assume a single-floor setting, and HOV-SG is restricted to the same floor for fair comparison. To examine whether the performance gain of our method stems from richer stored information alone or from the proposed fusion-based retrieval design, we additionally evaluate HOV-SG (w/ desc.), a variant in which HOV-SG receives the same instance-level captions generated by our captioning pipeline. Instance matching between HOV-SG predictions and our instances is performed using an IoU threshold of 0.5 to ensure that caption information is provided for the same physical objects.

\noindent \textbf{Query Types}: To assess robustness to query formulation, we define five query types that systematically vary the type and amount of information provided. (o) provides only the object category name, representing the simplest retrieval scenario. (o,r) additionally specifies the room in which the target is located. (o,d) combines the category name with a short visual description such as color or material. (o,r,d) provides category, room, and description together. (a,r,d) omits the explicit category name entirely and instead uses an abstract reference to the object's function or appearance, combined with room and description cues. This last type is the most challenging as it requires the system to infer the target category from indirect semantic cues.

\noindent \textbf{Metrics}: The object retrieval success rate (O-SR) measures whether the predicted instance matches the intended ground-truth instance. A prediction is counted as correct if the IoU with the matched ground-truth instance exceeds 0.30, following the threshold used in HOV-SG. For query types that do not uniquely specify a single instance, the criterion is relaxed. For (o), any ground-truth instance of the correct category is accepted. For (o,r), any ground-truth instance of the correct category in the correct room is accepted. In these cases, the predicted mask is matched to the ground-truth instance yielding the highest overlap, and the labels are then verified.

The navigation success rate (N-SR) measures whether the agent successfully reaches the target location from the assigned starting position. N-SR is computed over trials in which object retrieval succeeded, measuring navigation precision given correct instance identification. For VLMaps, which does not perform instance retrieval, N-SR is evaluated over all trials. A trial is counted as successful if the agent's final stopping position lies within one meter of the ground-truth instance location.

\begin{table}[t]
\centering
\caption{Object retrieval success rates (O-SR) and navigation success rates (N-SR) from language queries on the \ac{HM3DSEM} dataset. For each scene and for every query type, 50 trials are conducted using randomly sampled instances from the 30 most frequent object categories, yielding 250 trials per query type in total across five scenes. VLMaps is excluded from the object-retrieval comparison because its grid-wise feature storage does not maintain instance-level identity, making instance-based retrieval evaluation structurally infeasible. Navigation success rate (N-SR) is reported for VLMaps as positional goal-reaching remains evaluable independently of instance retrieval. Full+SPR denotes the variant that replaces MTEFR with a single-pass LLM inference over all instance information; the abbreviation is used for brevity in the table.}
\label{tab4}

\resizebox{\linewidth}{!}{%
\begin{tabular}{|l|c c|c c|c c|c c|c c|}
\hline
\multirow{3}{*}{\textbf{Method}}
& \multicolumn{10}{c|}{\textbf{Query type}} \\
\cline{2-11}
& \multicolumn{2}{c|}{(o)}
& \multicolumn{2}{c|}{(o,r)}
& \multicolumn{2}{c|}{(o,d)}
& \multicolumn{2}{c|}{(o,r,d)}
& \multicolumn{2}{c|}{(a,r,d)} \\
\cline{2-11}
& O-SR [\%] & N-SR [\%]
& O-SR [\%] & N-SR [\%]
& O-SR [\%] & N-SR [\%]
& O-SR [\%] & N-SR [\%]
& O-SR [\%] & N-SR [\%] \\
\hline
VLMaps & -- & 12.00 & -- & 6.00 & -- & 4.00 & -- & 8.00 & -- & 2.00 \\
HOV-SG & 40.00 & 15.00 & 8.00 & 18.18 & 10.00 & 23.07 & 8.00 & 18.18 & 2.00 & 0.00 \\
HOV-SG (w/ desc.) & 42.00 & 19.05 & 16.00 & 25.47 & 14.00 & 21.05 & 10.00 & 23.07 & 2.00 & 0.00 \\
\textbf{Ours (Full+MTEFR)} & \textbf{70.00} & \textbf{38.57} & \textbf{26.00} & 42.31 & \textbf{22.00} & 40.91 & \textbf{22.00} & 40.91 & \textbf{14.00} & 35.71 \\
\hline
Ours (Full+SPR) & 58.00 & 34.48 & \textbf{26.00} & 38.46 & 16.00 & 43.75 & 18.00 & \textbf{44.44} & 6.00 & \textbf{52.00} \\
Ours (Obj.) & 58.00 & 36.21 & 24.00 & 41.67 & 12.00 & \textbf{50.00} & 18.00 & \textbf{44.44} & 4.00 & 48.00 \\
Ours (Obj.+Desc.) & 60.00 & 35.00 & 24.00 & 45.83 & 16.00 & 43.75 & 18.00 & \textbf{44.44} & 6.00 & \textbf{52.00} \\
Ours (Obj.+Room) & 60.00 & 38.33 & \textbf{26.00} & \textbf{46.15} & 14.00 & 48.22 & 14.00 & 50.00 & 8.00 & 37.50 \\
\hline
\end{tabular}
}
\end{table}

\noindent \textbf{Results}: 
Table \ref{tab4} reports the quantitative results. On object retrieval, our method (Full+MTEFR) achieves the highest O-SR on average across all query types, with particularly large margins over HOV-SG on description-based and abstract queries. Across all methods, the object-only query type yields notably higher object retrieval success rates compared to other query types, as any instance of the correct category is accepted as a valid prediction. HOV-SG extracts only object and room keywords from the query, leaving descriptive content unused. This is directly confirmed by comparing the room-based and room-with-description query types, where adding descriptive attributes yields no improvement for HOV-SG, demonstrating that descriptive information is effectively discarded.

On navigation, our method (Full+MTEFR) achieves the highest conditional N-SR among all primary baselines across most query types. For VLMaps, N-SR is computed over all trials since instance-level retrieval is structurally infeasible. The resulting N-SR values are the lowest among all methods, reflecting the fundamental limitation that point-wise feature storage cannot identify specific target instances for precise goal-directed navigation. HOV-SG achieves moderate conditional N-SR, indicating that its navigation module can reach the target when 
retrieval succeeds. However, its low O-SR substantially limits the practical utility of this capability.

To isolate the contribution of MTEFR from that of the enriched instance representation, we compare against HOV-SG (w/ desc.), which receives identical instance-level captions without the fusion-based retrieval mechanism, and against the variant that replaces MTEFR with a single-pass LLM inference over all instance information (denoted Full+SPR in Table \ref{tab4} for brevity). HOV-SG (w/ desc.) shows only marginal improvement over the standard HOV-SG baseline despite receiving the same instance-level captions as our method. This result demonstrates that access to richer stored information alone is insufficient: without a mechanism to selectively leverage the relevant attributes for each query formulation, descriptive information cannot be effectively utilized. This confirms that the fusion-based query-adaptive retrieval design of the proposed MTEFR module, rather than the enriched instance representation alone, is the primary driver of performance improvement.

The ablation results reveal the individual contribution of each information type. Comparing Ours (Obj.) against Ours (Obj.+Room) shows improvement on (o,r) queries, confirming that room context contributes meaningfully to spatially-grounded retrieval. Comparing Ours (Obj.) against Ours (Obj.+Desc.) shows improvement on (o,d) queries, confirming the benefit of descriptive attributes for attribute-specific queries.Comparing Ours (Full+MTEFR) against the single-pass variant reveals the benefit of the hierarchical expert routing structure: while this single-pass variant provides all instance information to a single LLM pass in one inference step, MTEFR dynamically routes each query to type-specialized experts and integrates their outputs via score-level fusion, yielding consistent gains particularly on (o) and (a,r,d) queries where a single dominant information type or abstract intent drives the selection. The full model (Full+MTEFR) achieves the best overall performance, demonstrating that object category, room context, and descriptive attributes contribute complementarily and that their structured integration through expert routing is essential for robust performance across diverse query formulations. Among the ablation variants, conditional N-SR occasionally exceeds that of Full+MTEFR on 
individual query types, attributable to a small denominator effect: variants with low O-SR have fewer retrieval-successful trials, and a small number of 
navigation successes can yield a disproportionately high conditional rate. N-SR for ablation variants should therefore be interpreted alongside O-SR rather than in isolation.

\noindent \textbf{Failure Case Analysis}: 
The abstract expert LLM in MTEFR provides dedicated support for handling (a,r,d) queries, which is reflected in the substantially higher O-SR compared to HOV-SG (14.00\% vs. 2.00\%). However, the absolute success rate of 14.00\% remains low relative to the other query types (22.00\%–70.00\%), as shown in Table \ref{tab4}. Analyzing the failure cases reveals a recurring pattern. For instance, when the target is a cabinet in the living room and the query is given as "Find something for storage in the living room," the system incorrectly retrieves a table as the final result. The abstract expert LLM must infer the target solely through functional and semantic reasoning without any explicit categorical anchor. Although "something for storage" most strongly associates with cabinet, the abstract expert LLM also considers loosely connected candidates such as tables and shelves, and fails to sufficiently distinguish the intended target among them. This indicates that when an abstract expression admits an overly broad set of plausible semantic connections, pinpointing the correct instance becomes inherently challenging. Nevertheless, the proposed framework demonstrates the strongest capability in handling such queries among the compared methods.

Additionally, when an object is observed from only a limited set of viewpoints, the captioning module may produce inaccurate visual attribute descriptions, which can mislead the description expert during retrieval. This is a fundamental limitation inherent to any VLM-based captioning approach that operates on finite observations.

\begin{table}[t]
\small
\centering
\caption{Comparison of storage size. Results are measured in megabytes (MB) across five \ac{HM3DSEM} dataset scenes. For multi-floor scenes (00862, 00873, 00890), only the first floor is considered.}
\label{tab5}
\resizebox{0.6\columnwidth}{!}{
\begin{tabular}{|l l| c | c c c c c|}
\hline
\multirow{2}{*}{\textbf{Dim.}} &
\multirow{2}{*}{\textbf{Method}} &
\multirow{2}{*}{\textbf{Avg.}} &
\multicolumn{5}{c|}{Scene} \\
\cline{4-8}
& & & 00824 & 00829 & 00862 & 00873 & 00890 \\
\hline
3D & HOV-SG & 248 & 254 & 342 & 190 & 263 & 187 \\
2D & VLMaps & 118 & 186 & 99 & 179 & 68 & 57 \\
2.5D & \textbf{Ours} & \textbf{10} & \textbf{11} & \textbf{10} & \textbf{13} & \textbf{9} & \textbf{9}\\
\hline
\end{tabular}
}
\end{table}

\subsection{Storage Cost}\label{subsec:exp4}
Storage cost is an important consideration in mapping. We evaluated the storage requirements of VLMaps, HOV-SG, and our approach on five \ac{HM3DSEM} scenes, as reported in Table \ref{tab5}. Since both our method and VLMaps operate under a single-floor assumption, only the first floor was considered for multi-floor scenes (00862, 00873, 00890).

Despite being a 3D scene-graph model, HOV-SG required the largest storage (avg. 248 MB), as its volumetric representations grow substantially with scene complexity. Although VLMaps is 2D-based, it still required considerable storage (avg. 118 MB) due to its design of storing dense semantic features at every grid cell. In contrast, our method stores features at the object level within a compact 2.5D representation, reducing storage to an average of 10 MB, which is approximately 96\% less than HOV-SG and 92\% less than VLMaps. Notably, this reduction is achieved without discarding instance-level semantic information, highlighting the efficiency of the proposed object-centric 2.5D design.

\section{Conclusion}
\label{sec:conclusion}
This paper presents Instance-Enriched Semantic Maps, a unified framework integrating instance-level 2.5D mapping, room-level semantic segmentation, and instance captioning to support open-vocabulary navigation. The proposed 2.5D representation outperforms the 3D baseline by an improvement of over 27\% in average prediction-normalized AUC across datasets, including reliable preservation of small objects with substantially lower instance count gaps, while reducing storage by approximately 96\%. Room-level semantic segmentation achieves
consistent improvements over the \ac{CLIP} baseline across all metrics, with \ac{mIoU} increasing from 0.53 to 0.75, confirming that structural priors effectively suppress pixel-level noise and enforce spatially coherent region boundaries. Building on this enriched representation, the proposed MTEFR module achieves an average improvement of over 17\% in object retrieval and 23\% in navigation success compared to HOV-SG across diverse query types.

Despite these advances, limitations remain. The use of multiple vision-language models introduces computational overhead restricting real-time deployment, and room segmentation relies on physical partitions, reducing effectiveness in open-plan environments. Abstract queries admitting overly broad semantic interpretations remain inherently challenging for the retrieval stage.

Future work will explore lightweight model distillation, quantized on-device VLM inference, and asynchronous captioning pipelines. Additional directions include contextual room segmentation that infers spatial structure from object distributions rather than physical boundaries, and more constrained abstract reasoning through functional category hierarchies. Although real-world deployment is beyond the current experimental scope, the modular architecture is well-suited for adaptation to real robotic platforms, with each stage implementable as independent \ac{ROS} nodes and camera poses obtainable from visual-inertial \ac{SLAM} systems such as ORB-SLAM3 \citep{2021orbslam3} or RTAB-Map  \citep{2019rtab}. The compact 10MB map footprint further supports deployment on resource-constrained embedded platforms, and the secondary fusion pass already tolerates moderate pose drift through geometric and semantic similarity matching.

\appendix
\section{Prompts for Caption Generation}
\label{sec:appendixA}

\begin{figure}[b]
    \centering
    \includegraphics[width=0.58\textwidth]{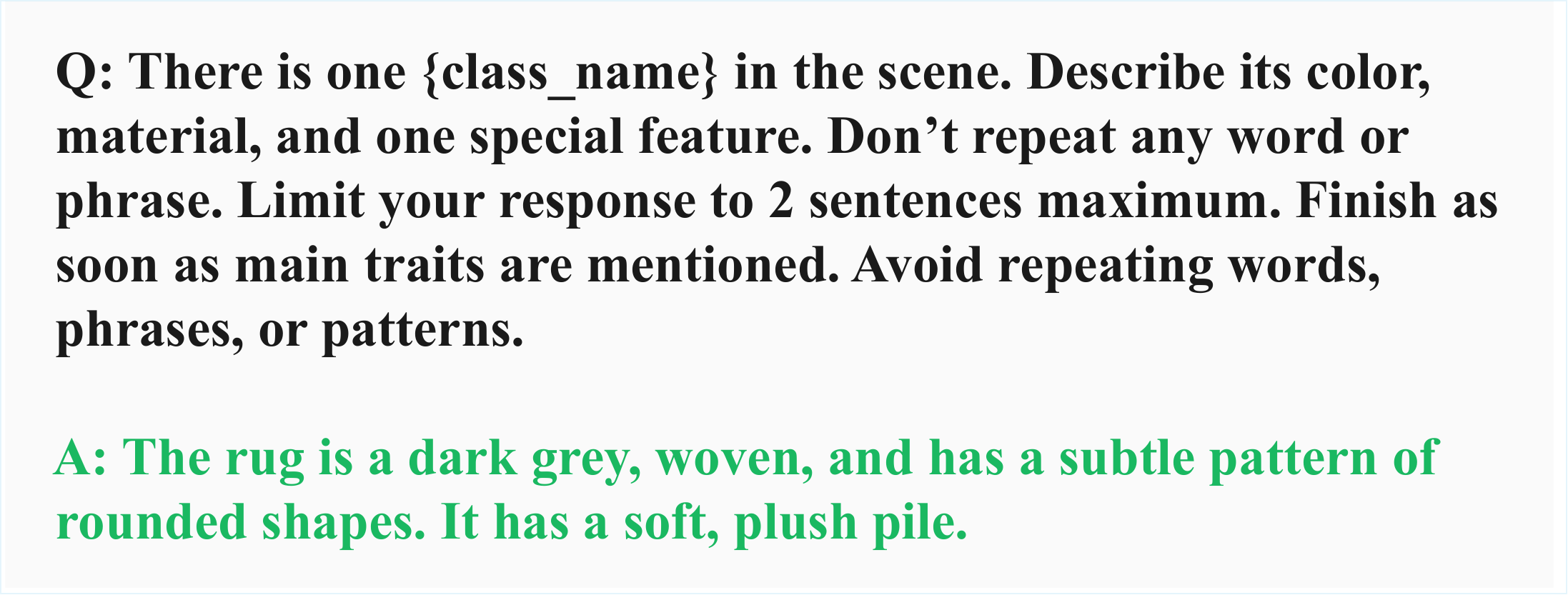}
    \caption{\textbf{Example of the prompt used for instance-level captioning with LLaMA 3.2 Vision.}}
    \label{fig8}
\end{figure}

\begin{figure}[t]
    \centering    \includegraphics[width=0.58\textwidth]{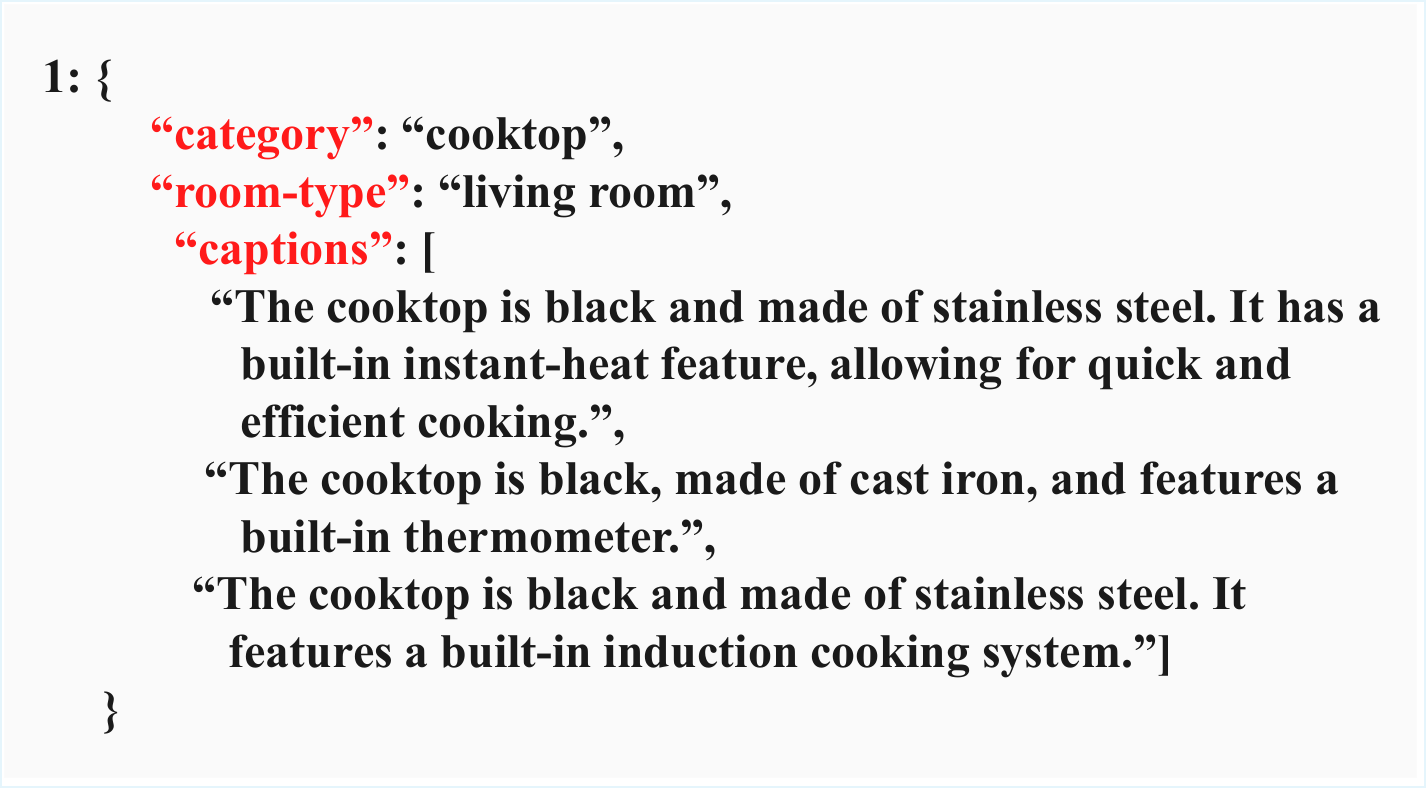}
    \caption{\textbf{Example of the result with LLaMA 3.2 Vision.}}
    \label{fig9}
\end{figure}

As presented in Sec. \ref{subsec:methodC}, using the VLM-based methods, we used the following prompts to derive object-related descriptions. Our pipeline leverages two complementary models for instance captioning: LLaMA 3.2 Vision for visual attribute extraction and GPT-4 for textual aggregation. While LLaMA 3.2 Vision provides rich object-level descriptions, it often introduces redundant wording and repeated expressions that hinder concise representation. To address this issue, we designed prompts instructing the model to eliminate unnecessary repetitions, and refine the language into a compact yet informative caption.

LLaMA 3.2 Vision was guided to produce descriptive captions for segmented instances, conditioned on representative keyframe images and the corresponding predicted category label. The employed prompt is shown in Fig. \ref{fig8}. The "user" content is shown in black and the "assistant" content in green. The resulting output is presented as depicted in Fig. \ref{fig9}.

\begin{figure}[t]
    \centering    \includegraphics[width=0.58\textwidth]{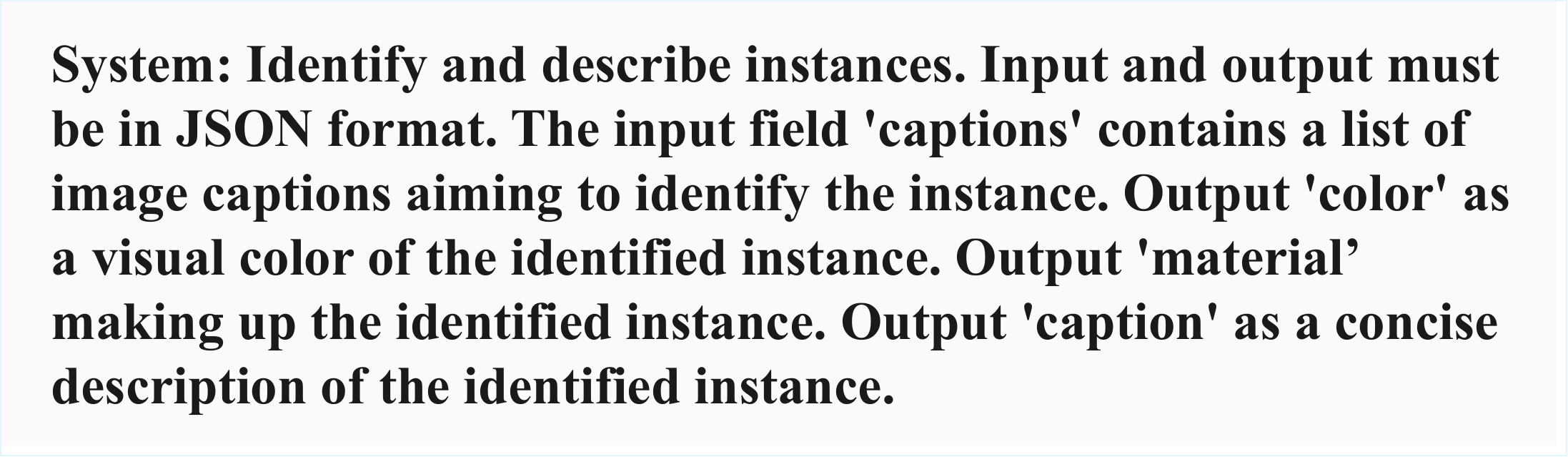}
    \caption{\textbf{Example of the system prompt with GPT-4.}}
    \label{fig10}
\end{figure}

GPT-4 is employed to refine instance-level captions generated by LLaMA 3.2 Vision. The model eliminates redundant expressions and extracts structured attributes for each instance. To guide this process, a tailored system prompt was used to explicitly instruct the model to summarize key visual traits into concise captions. The system prompt is shown in Fig.\ref{fig10}. 

As illustrated in Fig.\ref{fig3}, this refinement step produces the final instance representations consisting of a room tag, a compact natural language description, and structured key–value attributes, which collectively support robust open-vocabulary navigation.

\section{Prompts for Navigation}
\label{sec:appendixC}
\begin{figure}[t]
    \centering    \includegraphics[width=0.54\textwidth]{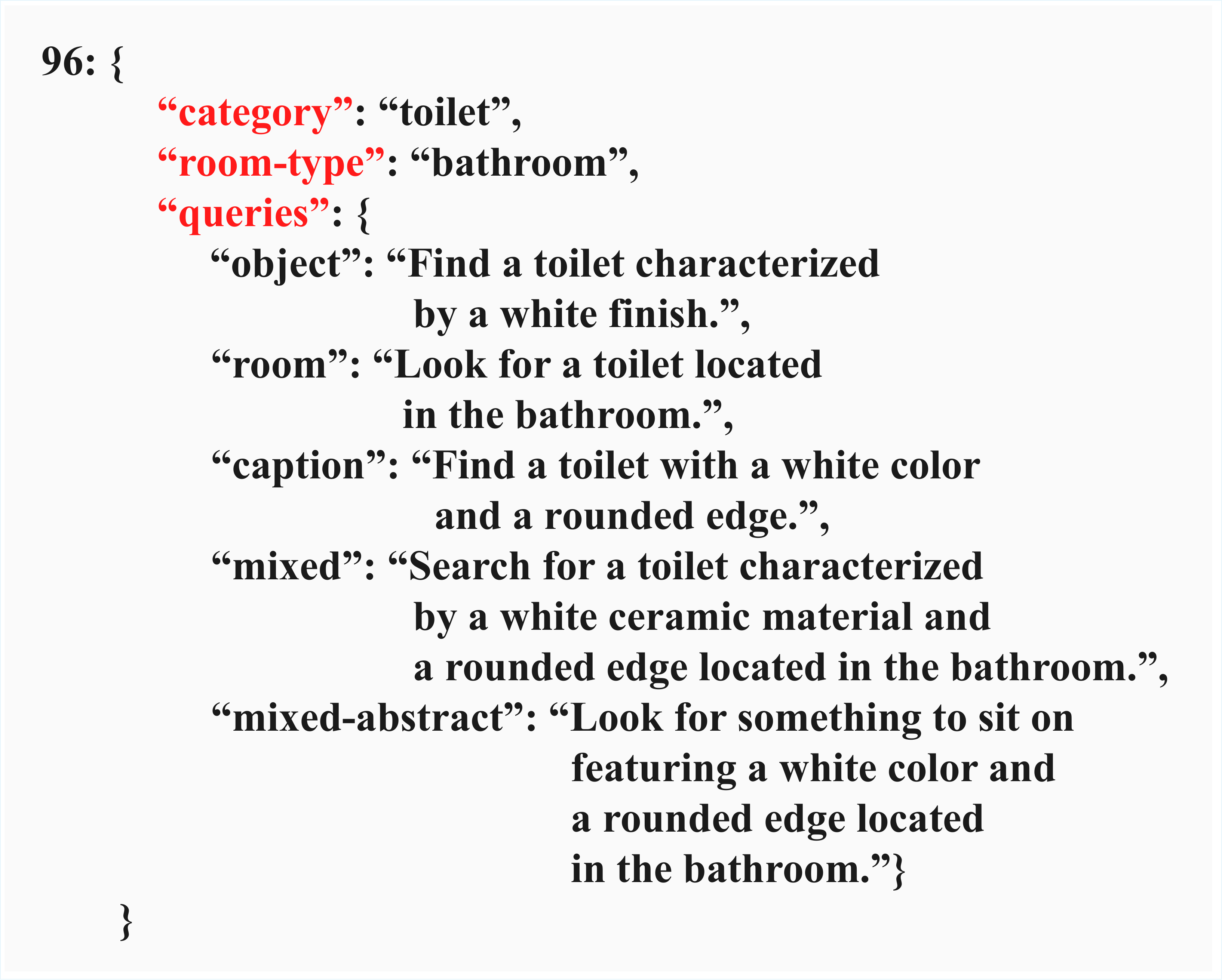}
    \caption{\textbf{Example of a navigation query.}}
    \label{fig11}
\end{figure}

To evaluate navigation under diverse language conditions, we automatically generated queries using GPT-based prompting. Each query was conditioned on three inputs: a representative frame of the target instance, its instance-level category label, and the corresponding room category label. We defined five query types:

1) Object-based query: explicitly referring to the target object category

2) Room-based query: emphasizing the room context in which the object is located

3) Caption-based query: leveraging descriptive attributes extracted from instance-level captions

4) Mixed query: combining object, room, and caption information into a single instruction

5) Abstract query: indirect expressions of the object (e.g., "something to sit on") instead of explicit category names

Predefined lexical variants were employed to generate queries with various linguistic expressions.
This design ensures that queries remain both diverse 
and systematically comparable across conditions. An example output is shown in Fig. \ref{fig11}, which illustrates the five types of queries.

\bibliographystyle{elsarticle-harv}
\bibliography{packages_eaai/references}

\end{document}